\documentclass[lettersize,journal]{IEEEtran}
\usepackage{amsmath,amsfonts}
\usepackage{algorithmic}
\usepackage{algorithm}
\usepackage{array}
\usepackage[caption=false,font=normalsize,labelfont=sf,textfont=sf]{subfig}
\usepackage{textcomp}
\usepackage{stfloats}
\usepackage{url}
\usepackage{verbatim}
\usepackage{graphicx}
\usepackage{cite}
\newif\ifarxiv

\arxivtrue       % arXiv version
\ifarxiv
    \usepackage[font=small]{caption}
\fi

\usepackage{etoolbox}
\usepackage[utf8]{inputenc} % allow utf-8 input
\usepackage[T1]{fontenc}    % use 8-bit T1 fonts
\usepackage{hyperref}       % hyperlinks
\usepackage{url}            % simple URL typesetting
\usepackage{booktabs}       % professional-quality tables
\usepackage{amsfonts}       % blackboard math symbols
\usepackage{nicefrac}       % compact symbols for 1/2, etc.
\usepackage{microtype}      % microtypography
\PassOptionsToPackage{table}{xcolor}
\usepackage{xcolor}         % colors

\usepackage{cuted}
\usepackage{capt-of}

\usepackage{graphicx}
\usepackage{mathtools}
\usepackage{tabularx} 
\usepackage{wrapfig}
\usepackage{multirow}
\usepackage{bm}
\usepackage{amsmath}
\usepackage{amssymb}
\usepackage{arydshln}
\usepackage{epsfig}
\usepackage{adjustbox}
\usepackage{lipsum}
\usepackage{afterpage}
\usepackage[most]{tcolorbox}
\usepackage{float}
\usepackage{stfloats}
\usepackage{algorithm}
\usepackage{algorithmic}

\definecolor{hlgreen}{HTML}{84994F}
\newcommand{\best}[1]{\cellcolor{hlgreen!35}\textbf{#1}}
\newcommand{\sbest}[1]{\cellcolor{hlgreen!15}#1}
\definecolor{improvementcolor}{HTML}{84994F}

\tcbset{
  mypromptstyle/.style={
    enhanced,
    breakable,
    sharp corners,
    colback=gray!5!white,
    colframe=black,
    boxrule=0.5pt,
    width=\textwidth,
    boxsep=0pt,  % no padding
    left=0pt,
    right=0pt,
    top=0pt,
    bottom=0pt,
  }
}

\definecolor{foggyblue}{RGB}{87, 123, 193}
\newcommand{\prompttitle}[1]{%
  \noindent
  \colorbox{foggyblue}{%
    \makebox[\dimexpr\linewidth-2\fboxsep\relax][l]{%
      \textcolor{white}{\bfseries #1}
    }%
  }%
  \par\vspace{0.6em}
}

\makeatletter
\renewcommand{\@fnsymbol}[1]{\ifcase#1\or\dag\else\@arabic{#1}\fi}
\makeatother

\begin{document}

\ifarxiv
    \title{%
    {\LARGE\bfseries
    \makebox[\textwidth][c]{%
      \makebox[1.04\textwidth][c]{%
        CycleVLA: Proactive Self-Correcting Vision-Language-Action Models%
      }%
    }\\[-0.20em]
    \makebox[\textwidth][c]{%
      \makebox[1.04\textwidth][c]{%
        via Subtask Backtracking and Minimum Bayes Risk Decoding%
      }%
    }%
    }}
\else
    \title{CycleVLA: Proactive Self-Correcting Vision-Language-Action Models via
    Subtask Backtracking and Minimum Bayes Risk Decoding}
\fi

\makeatletter
\patchcmd{\@maketitle}
  {\vskip1.0em\par}
  {\vskip2.0em\par}
  {}
  {}
\makeatother

\ifarxiv
    \author{Chenyang Ma$^{1}$, Kai Lu$^{2}$, Guangyu Yang$^{3\dagger}$, Jiuming Liu$^{3}$, Shitong Xu$^{1}$,\\ Bill Byrne$^{3}$, Ioannis Havoutis$^{2}$, Niki Trigoni$^{1}$, Andrew Markham$^{1}$
    \thanks{$^{1}$Department of Computer Science, University of Oxford}%
    \thanks{$^{2}$Oxford Robotics Institute, University of Oxford}%
    \thanks{$^{3}$Department of Engineering, University of Cambridge}%
    \thanks{$^{\dagger}$Corresponding author.}%
    }
\else
  \author{Anonymous Authors}
\fi
    

% % column for its text to clear the IEEEpubid mark.

\maketitle

%------------------------------------------------------------------
\begin{strip}
\ifarxiv
    \vspace{-6em}
\else
    \vspace{-3em}
\fi
  \centering
  \footnotesize
  \setlength{\abovecaptionskip}{0.1cm}
  \includegraphics[width=1\linewidth]{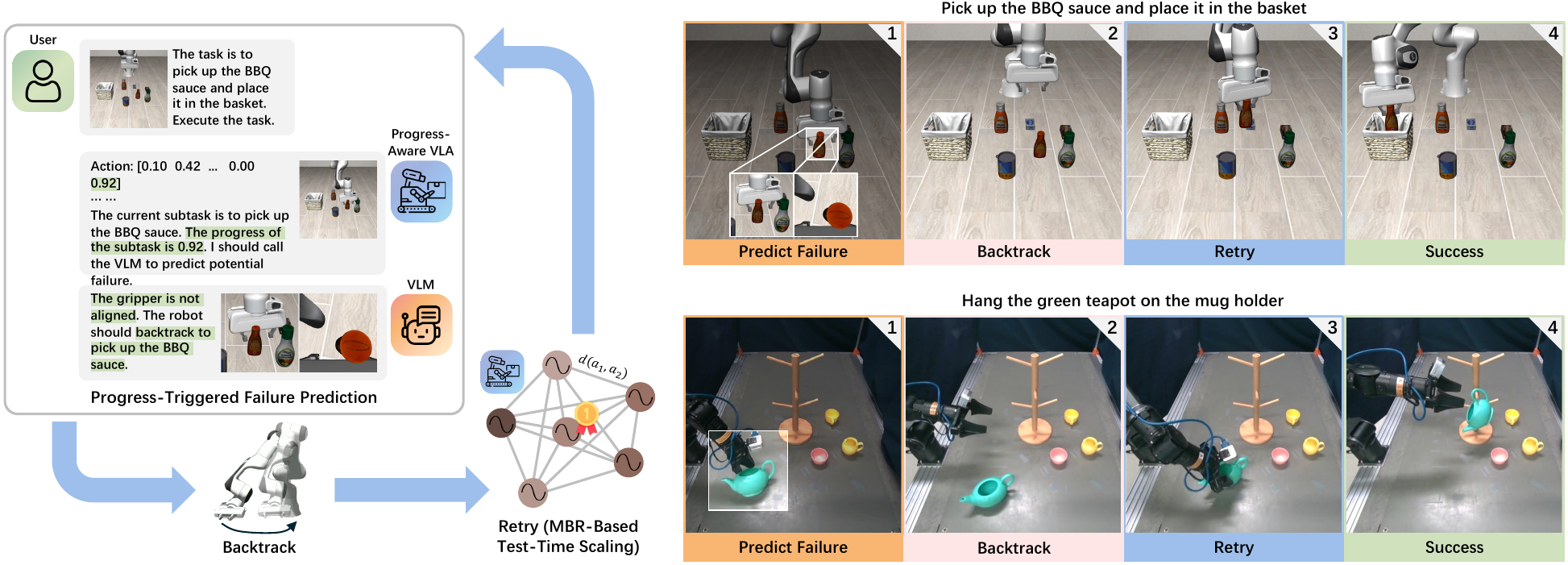}
  \vspace{-1.5mm}
  \captionof{figure}{\textbf{Vision–Language–Action models with proactive self-correction capabilities.} We introduce \textbf{CycleVLA}, which enables VLAs to anticipate incipient failures and recover before execution collapses. CycleVLA first augments a VLA to estimate subtask-level progress and flag critical subtask transition points, where failures most frequently occur. Then, at these points during inference, a VLM is queried to predict whether the current execution will fail and to decide whether to backtrack. Upon backtracking, the VLA retries using test-time scaling via Minimum Bayes Risk decoding to improve success. This cycle repeats until the task succeeds or execution terminates.}
  \label{fig:teaser}
  % \vspace{-1.0em}
\end{strip}

\begin{abstract}

Current work on robot failure detection and correction typically operates in a post hoc manner, analyzing errors and applying corrections only after failures occur. This work introduces \textbf{CycleVLA}, a system that equips Vision-Language-Action models (VLAs) with proactive self-correction, the capability to anticipate incipient failures and recover before they fully manifest during execution. CycleVLA achieves this by integrating a progress-aware VLA that flags critical subtask transition points where failures most frequently occur, a VLM-based failure predictor and planner that triggers subtask backtracking upon predicted failure, and a test-time scaling strategy based on Minimum Bayes Risk (MBR) decoding to improve retry success after backtracking. Extensive experiments on the LIBERO and LIBERO-Plus simulation benchmarks show that CycleVLA surpasses the state-of-the-art VLA $\pi_{0.5}$, improves success rates by correcting execution failures across VLAs of varying capability, from under-trained ones to fully converged policies, and that MBR serves as an effective zero-shot test-time scaling strategy for VLAs. On a real robot, CycleVLA reaches a 91\% average success rate on one precise and two long-horizon manipulation tasks. We further conduct stress tests with multiple manually injected perturbations (e.g., swapping in a distractor at the expected location while relocating the true target object mid-execution), where CycleVLA corrects $\sim$80\% of injected failures and maintains success rates comparable to unperturbed execution.
\end{abstract}

%------------------------------------------------------------------
\ifarxiv
\else
    \begin{IEEEkeywords}
    Vision-language-action models, Failure recovery, Test-time scaling
    \end{IEEEkeywords}
\fi

%------------------------------------------------------------------
\section{Introduction}
Humans constantly adjust their actions when they sense something going wrong, such as tightening their grip when a glass begins to slip, or adjusting the steering wheel before drifting out of a lane. These corrections happen during the act, not after, because a shattered glass cannot be recovered, and a car off the road is already in danger. Once an error has fully occurred, the opportunity for correction has passed.

We call this \textit{proactive self-correction}~\cite{XieTSF22}, the ability to detect incipient errors during execution and adapt before failure fully manifests. Replicating this ability in robot agents remains an open challenge. For instance, such a robot should recognize that a grasp is misaligned and reposition its gripper before knocking the object over, completing the task within the execution episode without external intervention. This stands in contrast to most recent work on robot failure detection and correction, which operates in a post hoc manner: execution failures are identified only as or after they occur~\cite{FaridSRM22, DuKDRLHFC23, GokmenHK23, AgiaSYCA0B24, ask_act_ram, collabvla_sun, SAFE_Gu}, and corrective~actions are applied retrospectively, often as residual policies~\cite{LiuBS23, DuanPKWTYKFMG25, lu2025kitchenvla, star_sakib}.

In this work, we aim to equip generalist robot foundation models, specifically Vision-Language-Action models (VLAs)~\cite{KimPKXB0RFSVKBT24, pi_levine, gr00t_fan}, with proactive self-correcting capabilities, as shown in Fig.~\ref{fig:teaser}. Our approach, \textbf{CycleVLA}, is motivated by the observation that many robot task failures occur at subtask transitions~\cite{MyersZMFL24, pi0_5_pi, ShiIEKPVTWWFLDG25, seqvla_yang, fan2025longvla}, and that progress near subtask completion provides strong cues for anticipating such failures (e.g., one can tell a peg is misaligned before it jams during insertion). Based on this insight, CycleVLA first introduces a finetuning pipeline that endows VLAs with explicit awareness of subtask progress, addressing a key limitation of existing VLAs that lack mechanisms for stopping or progress estimation~\cite{seqvla_yang}. We achieve this by decomposing demonstrations into aligned subtasks and finetuning the VLA with extended action dimensions that predict both a stop signal and subtask progress.

With a progress-aware VLA, the next challenge is deciding whether to transition to the next subtask or intervene near subtask termination, before failure occurs. CycleVLA addresses this through an explicit backtracking mechanism~\cite{KeLBHGLGCS19, Shi2025SmartWayEW, err_du} guided by a Vision-Language Model (VLM)~\cite{llama3_Dubey, BrownMRSKDNSSAA20, ma2024spatialpin, yang2022touch, LiuLLL24, liang2026cairn}. At test time, the VLM acts as a zero-shot failure predictor and planner: if no failure is predicted, execution continues; otherwise, the system reverts to the earliest subtask that restores missing preconditions and retries.

Finally, to improve the success of retries after backtracking, CycleVLA adopts a zero-shot test-time scaling strategy based on Minimum Bayes Risk (MBR) decoding~\cite{KumarB04}. Since most VLAs are trained via imitation learning~\cite{ONeillRMGPLPGMJ24, KimPKXB0RFSVKBT24, LiuWLTCWX0025}, successful behaviors tend to cluster in high-density regions of the policy output space~\cite{kwok25a, rover_dai}, which makes consensus selection over multiple samples more likely to succeed. Inspired by recent applications of MBR decoding in large language models (LLMs) at inference time~\cite{EikemaA22, tie_chen, YangCLB24}, we sample multiple trajectories from the VLA and select the one that minimizes expected risk under a distance metric.

In summary, our main contributions are as follows:
% \vspace{-0.5em}
\begin{itemize}
\renewcommand\labelitemi{\scalebox{0.75}{$\bullet$}}
\item We introduce CycleVLA, a system that enables proactive self-correction in VLAs by combining (a) a finetuning pipeline that extends VLAs with subtask progress prediction, (b) a VLM-based failure predictor and planner that uses progress cues to decide if to transit or backtrack between subtasks, and (c) an MBR-based test-time scaling strategy that improves recovery after backtracking.
\item We demonstrate that MBR decoding serves as a zero-shot test-time scaling strategy for improving the success rate of VLA policies.
\item Extensive experiments on LIBERO and LIBERO-Plus simulation benchmarks show that CycleVLA improves success rates by correcting execution failures across VLAs of varying capability, surpassing the state-of-the-art $\pi_{0.5}$, and remaining effective on under-trained VLAs.
\item We validate CycleVLA on a real robot, reaching a 91\% average success rate on one precise and two long-horizon manipulation tasks, and correcting $\sim$80\% of failures under stress tests with multiple manually injected perturbations.
\end{itemize}

%------------------------------------------------------------------
\section{Related Work}

\subsection{Robot Failure Detection and Correction}
Most prior work addresses robot failure detection in isolation. Given the scarcity of robot failure data, many approaches rely on failures observed at inference time, using conformal prediction~\cite{FaridSRM22, AgiaSYCA0B24, no_fail_data_xu, SAFE_Gu, yang2022sparse}, state-based anomaly detection~\cite{WongTKM0SM21, XieTSF22, GokmenHK23, LiuDMZ24}, or LLMs/VLMs for failure recognition~\cite{DuKDRLHFC23, SinhaEAFS024, GuoWZC24, AgiaSYCA0B24, li2025selfcorrecting}. Another line of work studies post hoc failure correction, applying retrospective analysis and residual recovery policies after failures occur~\cite{LiuBS23, DuanPKWTYKFMG25, lu2025kitchenvla, ma2025coopera, star_sakib, he2026pact}. More recent work explores proactive correction during execution, such as combining visuomotor policies with visual world models to anticipate future states~\cite{LiuZBZLDZ24, ZhaoLKFZWLMHFHL25, ref_plan_feng, vjepa2_assran, tu2026unitac}, but these incur substantial overhead and architectural changes. In contrast, our work equips generalist robot foundation models with proactive self-correction via explicit VLM-based failure prediction, subtask-level backtracking~\cite{KeLBHGLGCS19, Shi2025SmartWayEW, err_du}, and retry. The closest works are PAINT~\cite{XieTSF22} and Bellman-Guided Retrials~\cite{err_du}, but PAINT requires human intervention and the latter does not target generalist policies.

% Other approaches rely on human-in-the-loop supervision, with humans monitoring execution and intervening to provide direct feedback~\cite{ShiHZSPLLF24, TabatabaeiKJ25, ma2025coopera}, or assisting the robot in correcting predicted errors when failures are detected and flagged by the system~\cite{FaridSRM22, ask_act_ram, GokmenHK23, collabvla_sun, LiuZBZLDZ24}.

\subsection{Data Augmentation for VLAs}
VLA training lacks a standard data augmentation strategy, and recent work explores several directions. To tackle long-horizon tasks, some approaches decompose demonstrations into subtasks~\cite{MyersZMFL24, pi0_5_pi, ShiIEKPVTWWFLDG25, seqvla_yang, dexvla_wen, fan2025longvla, rdd_yan, ma2026furniturevla}, with the active subtask inferred at inference by an external system~\cite{ShiIEKPVTWWFLDG25} or the VLA itself~\cite{ZawalskiCPMFL24, pi0_5_pi}. Some works incorporate textual reasoning into training to enable explicit reasoning at inference~\cite{ZawalskiCPMFL24, chatvla_zhou, chatvla2_zhou, ecotlight_chen}. Visual augmentation has also been explored: Cosmos~\cite{nvidia2025cosmos} applies video style transfer, while others highlight task-relevant objects~\cite{HancockRM25, iavla_hannus}, akin to visual prompting in VLMs~\cite{BarGDGE22}. Our work also adopts subtask decomposition, but specifically focuses on teaching VLAs when to stop at subtask transitions and to track their own progress—capabilities largely absent from existing VLAs~\cite{seqvla_yang}.

\subsection{Test-Time Scaling for VLAs}
Test-time scaling has proven effective for LLMs~\cite{snell_scale_llm, deepseek_r1, GPT-o3_openai}, and recent work explores its application to VLAs. RoboMonkey~\cite{kwok25a} samples actions with Gaussian perturbations and uses a trained VLM to select among candidates. Rover~\cite{rover_dai} and V-GPS~\cite{NakamotoMKL24} score sampled actions using learned reward or value models. In contrast, our MBR decoding~\cite{KumarB04}, inspired by its use in LLMs~\cite{EikemaA22, tie_chen, YangCLB24}, performs training-free consensus selection without external verifiers.

% MG-Select~\cite{verifier_free_jang} uses KL divergence from a reference action token distribution as a confidence metric for selecting the optimal action from multiple candidates. 

%------------------------------------------------------------------
\section{Preliminaries}\label{sec:Preliminaries}
We consider sequential decision-making with the current observation $o_t$ (e.g., RGB images and proprioception) and a natural language goal $g$. A VLA policy $\pi_\theta$ maps $(o_t, g)$ to robot actions. We adopt a continuous end-effector delta action representation
$a_t = [\Delta x_t, \Delta y_t, \Delta z_t, \Delta u_t, \Delta v_t, \Delta w_t, \gamma_t]^\top \in \mathbb{R}^7$,
where $(\Delta x_t, \Delta y_t, \Delta z_t) \in \mathbb{R}^3$ and $(\Delta u_t, \Delta v_t, \Delta w_t) \in \mathbb{R}^3$ denote translational and rotational displacements, and $\gamma_t \in \{0,1\}$ indicates the gripper's open/close state. Unlike autoregressive formulations, we assume the VLA employs parallel decoding to produce a continuous chunk of future actions in a single forward pass, i.e., $\pi_\theta(a_{t:t+H-1} \mid o_t, g)$ for chunk size $H$. At inference time, we further assume the policy supports stochastic decoding, enabling the generation of a finite set of action sequence hypotheses $\mathcal{A} = \{a^{(1)}_{t:t+H-1}, \dots, a^{(N)}_{t:t+H-1}\}$ via repeated forward passes (e.g., stochastic latent variables, diffusion-based noise sampling). 

% Additionally, we assume access to a dataset expert robot demonstrations, where each trajectory is paired with a natural language instruction.

% Additionally, we assume access to a dataset of expert demonstrations $\mathcal{D} = \{(\tau_i, I_i)\}_{i=1}^{N_D}$, where each trajectory $\tau_i$ consists of state-action sequences paired with a language instruction.

%------------------------------------------------------------------
\section{CycleVLA}\label{sec:CycleVLA}
\begin{figure*}[t]
  \centering
  \footnotesize
  \setlength{\abovecaptionskip}{0.1cm}
  \includegraphics[width=1\linewidth]{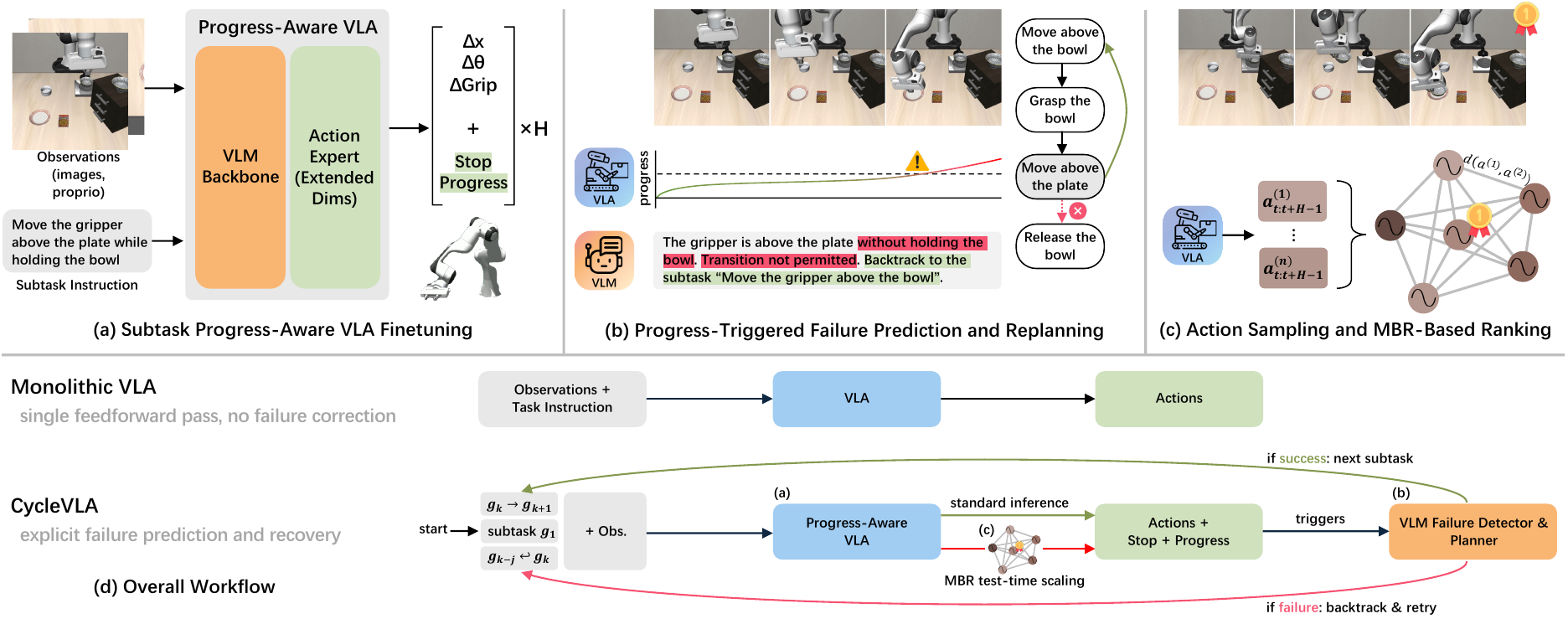}
  \vspace{-3mm} \caption{\textbf{CycleVLA.} (a) A finetuning pipeline that equips a VLA with subtask-level stop and progress prediction via extended action expert dimension and augmented subtask-decomposed training data. (b) At inference, predicted progress triggers a VLM-based failure predictor and planner, which decides whether to transit to the next subtask or backtrack, and selects the subtask to backtrack to. (c) After backtracking, the VLA retries execution using test-time scaling via MBR decoding to improve success. (d) The workflow of CycleVLA compared to a monolithic VLA without failure correction. See Appendices~\ref{sec:Implementation Details of MBR Decoding} and~\ref{sec:Prompt Details of Failure Predictor and Planner} for MBR implementation details and VLM exact prompts.}
  \label{fig:pipeline}
  \vspace{-1.0em} 
\end{figure*}

Our goal is to equip VLAs with proactive self-correction capabilities. Our approach, CycleVLA (Fig.~\ref{fig:pipeline}), builds on the observation that task failures often occur at subtask transitions, where progress near completion provides strong cues for anticipating failure. We first introduce a finetuning pipeline that constructs a subtask-decomposed dataset from demonstrations and trains VLAs to explicitly predict stop and progress signals with lightweight modifications (Sec.~\ref{sec:Learning Stop and Progress Signals for Subtask Execution}). At inference time, we use an off-the-shelf VLM as a failure predictor and planner to decide whether to transition or backtrack at subtask boundaries (Sec.~\ref{sec:Test-Time Scaling via Subtask Backtrack and MBR Decode}). Finally, we apply MBR decoding as a zero-shot test-time scaling strategy to improve retry success after backtracking (Sec.~\ref{sec:Test-Time Scaling via Subtask Backtrack and MBR Decode}).

\subsection{Learning Stop and Progress Signals for Subtask Execution}\label{sec:Learning Stop and Progress Signals for Subtask Execution}

\begin{figure}[t]
  \centering
  \footnotesize
  \setlength{\abovecaptionskip}{0.1cm}
  \includegraphics[width=1\linewidth]{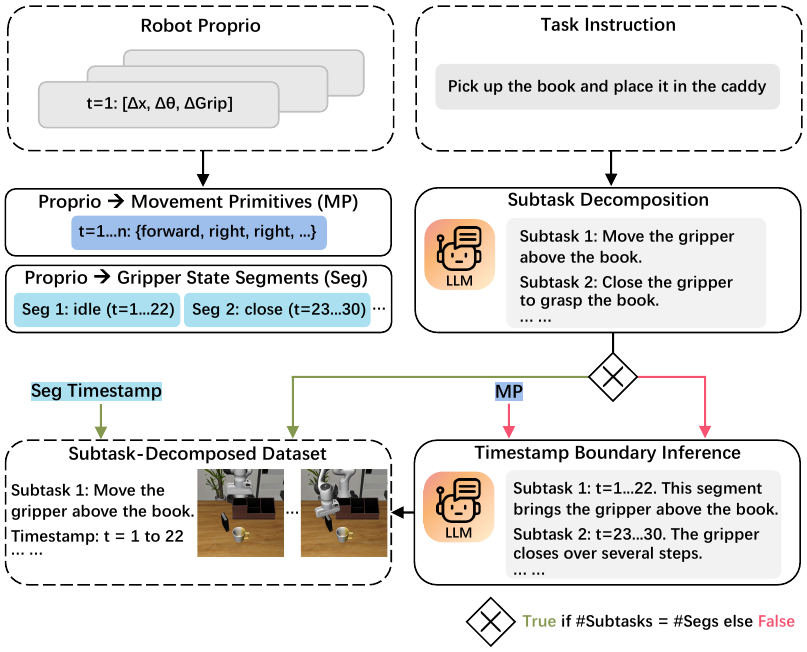}
  \vspace{-3mm} \caption{\textbf{Pipeline for constructing the subtask-decomposed dataset.} Following LLM subtask decomposition and extraction of movement primitives and gripper state segments, subtasks are directly aligned to gripper state segment timestamps when their counts match; otherwise, an LLM infers subtask boundaries from the movement primitive sequence. Please see Appendices~\ref{sec:Details and Evaluation of Subtask Decomposition} and~\ref{sec:Prompt Details of Subtask Decomposition} for implementation details and a human evaluation of subtask decomposition quality.}
  \label{fig:preprocessing_pipeline}
  \vspace{-1.0em} 
\end{figure}

\noindent\textbf{Constructing a Subtask-Decomposed Dataset.} Most robot demonstration datasets pair each trajectory with only a high-level task instruction, without subtask labels or their timestamps (e.g., LIBERO~\cite{LiuZGFLZS23} and RLBench~\cite{james2019rlbench}, with few recent exceptions such as BEHAVIOR-1K~\cite{li2024behavior1k}). We introduce a pipeline that uses LLMs to decompose demonstrations into subtasks with precise start/end timestamps and language instructions (Fig.~\ref{fig:preprocessing_pipeline}), inspired by prior work~\cite{MyersZMFL24, ZawalskiCPMFL24}.

Given a task instruction $g$, we first prompt an LLM to decompose it into a minimal sequence of atomic subtasks $(g_1,\ldots,g_K)$ using a constrained action vocabulary (e.g., move, rotate, open, close). In parallel, we extract per-step gripper states (open, close, or idle) and movement primitives (e.g., move forward, rotate clockwise, stop)~\cite{ZawalskiCPMFL24} from robot proprio by computing state differences over a sliding window. 

We then align subtasks to the trajectory using gripper state transition segments, which provide reliable subtask boundaries in manipulation tasks (e.g., continuous open/close segments correspond to grasping or releasing objects, while idle segments indicate periods of robot translational or rotational motion with no gripper actuation). If the number of LLM-proposed subtasks matches the number of gripper state segments, we directly pair each subtask with the corresponding segment timestamps. Otherwise, we prompt an LLM with the extracted movement primitive sequence (downsampled to a fixed maximum length to reduce context length) to infer subtask timestamp boundaries, enforcing continuous assignment without gaps while filtering spurious stop or inconsistent primitives. 

% For example, given "put the white mug on the left plate," the LLM outputs: \texttt{["Move the gripper above the white mug.", "Close the gripper to grasp the white mug.", "Move the gripper above the left plate.", "Open the gripper to release the white mug."]}

\vspace{0.5em}
\noindent\textbf{Subtask Finetuning with Extended Action Dims.} With the subtask-decomposed dataset, we finetune the VLA with extended action dimensions to predict stop and progress signals—capabilities not explicitly modeled in existing VLAs~\cite{seqvla_yang}. We extend the 7-dim action $a_t$ to 9-dim: $a_t = [\Delta x_t, \Delta y_t, \Delta z_t, \Delta u_t, \Delta v_t, \Delta w_t, \gamma_t, s_t, p_t]^\top \in \mathbb{R}^9$, where $s_t \in \{0,1\}$ denotes a stop signal indicating subtask termination, and $p_t \in [0,1]$ denotes subtask progress, discretized into bins of 0.1 based on the normalized timestep within each subtask. We explicitly separate these two signals because the stop signal must be precise to support correct subtask transitions, whereas the progress signal only needs to indicate proximity to subtask completion. Following NaVILA~\cite{cheng2024navila}, we oversample the last action step of each subtask during training to emphasize termination detection. At inference time, we binarize $s_t$ via thresholding, as with the gripper signal $\gamma_t$.

We predict stop and progress jointly with end-effector delta actions as scalar outputs, rather than introducing separate classification heads. This design aligns with the continuous nature of the VLA action space: end-effector displacements are floating-point values, and the gripper signal $\gamma_t$ is a bounded scalar. Predicting $s_t$ and $p_t$ in the same output space requires no architectural changes beyond widening the action dimension.

\subsection{Test-Time Scaling via Subtask Backtrack and MBR Decoding}\label{sec:Test-Time Scaling via Subtask Backtrack and MBR Decode}

\noindent\textbf{VLM as a Failure Predictor and Planner.} With a progress-aware VLA, we use an off-the-shelf VLM to predict failure and plan recovery at subtask boundaries. When the VLA-predicted progress reaches a threshold $\tau_p \in [0,1]$, we query the VLM with synchronized third-person and wrist-mounted camera views, the current subtask, and the subtask list. The VLM outputs a decision: transit to the next subtask, or backtrack to the earliest subtask that restores missing preconditions (e.g., if the grasped object drops midway, backtrack to the grasping subtask). We instruct the VLM to fuse global context from the third-person view (e.g., object identity, gripper pose, if the correct subtask is executed) with fine-grained cues from the wrist view (e.g., gripper alignment with the object, contact quality) via Chain-of-Thought reasoning~\cite{Wei0SBIXCLZ22} to decide.

When backtracking is triggered, we restore the robot configuration to the beginning of the target subtask by reverse-executing recorded delta actions~\cite{err_du}. Note that backtracking does not restore the object states; instead, the retried execution relies on the VLA re-perceiving the current scene, so CycleVLA works in real settings where the environment does not reset and errors, once they occur, must be corrected.

\vspace{0.5em}
\noindent\textbf{Sampling Actions and Ranking via MBR.} After backtracking restores a valid subtask precondition, the VLA retries the subtask from the same robot configuration. We apply test-time scaling by sampling multiple action chunk hypotheses and selecting a consensus one via MBR decoding~\cite{KumarB04}. From current observation $o_t$ and subtask $g_k$, we draw $N$ hypotheses $\mathcal{A}=\{a^{(1)}_{t:t+H-1},\dots,a^{(N)}_{t:t+H-1}\}$ from the stochastic policy $\pi_\theta(\cdot\mid o_t,g_k)$.

% MBR selects the decision that maximizes expected utility under model uncertainty. Let $u(y,h)$ denote the utility of choosing action chunk $h$ when $y$ is the ideal outcome. MBR chooses:
MBR selects the hypothesis that minimizes the expected risk under the policy distribution. Let $d(\cdot, \cdot)$ denote the distance between two action chunks. MBR chooses:
% \begin{equation}
% h^{\mathrm{MBR}} = \arg\max_{h \in \mathcal{Y}} \; \mathbb{E}_{Y \sim \pi_\theta(\cdot \mid o_t, g_k)} \!\left[u(Y,h)\right],
% \label{eq:1}
% \end{equation}
% where $\mathcal{Y}$ denotes the space of $H$-step action chunks. We approximate the expectation via Monte Carlo and restrict hypotheses to the sampled set $\mathcal{A}$:
% \begin{equation}
% a^{\mathrm{MBR}} = \operatorname*{argmin}_{a \in \mathcal{A}} \; \mathbb{E}_{a' \sim \pi_\theta(\cdot \mid o_t, g_k)} \!\left[d(a,a')\right],
% \label{eq:1}
% \end{equation}
% where we omit the time-step notation for simplicity. 
\begin{equation}
a^{\mathrm{MBR}}_{t:t+H-1} = \operatorname*{argmin}_{a_{t:t+H-1} \in \mathcal{A}} \; \mathbb{E}_{a'_{t:t+H-1} \sim \pi_\theta} \!\left[d(a_{t:t+H-1},a'_{t:t+H-1})\right].
\label{eq:1}
\end{equation}

We approximate the expectation via Monte Carlo using the hypothesis set $\mathcal{A}$ sampled from the policy $\pi_\theta(\cdot \mid o_t, g_k)$:
% \begin{equation}
% \hat{\mu}_u(h) = \frac{1}{N} \sum_{n=1}^{N} u\!\left(a^{(n)}_{t:t+H-1}, h\right),
% \label{eq:2}
% \end{equation}
% yielding the sampling-based $N$-by-$N$ MBR objective~\cite{EikemaA22}:
\begin{equation}
\mathcal{L}(a^{(i)}_{t:t+H-1}) = \frac{1}{N} \sum_{j=1}^{N} d\!\left(a^{(i)}_{t:t+H-1}, a^{(j)}_{t:t+H-1}\right),
\label{eq:2}
\end{equation}
where $\mathcal{L}$ is the estimated Bayes risk of an action chunk, yielding the sampling-based $N$-by-$N$ MBR objective~\cite{EikemaA22}:
% \begin{equation}
% a^{\mathrm{MBR}}_{t:t+H-1}
% = \arg\max_{h \in \mathcal{A}} \; \hat{\mu}_u(h).
% \label{eq:3}
% \end{equation}
\begin{equation}
a^{\mathrm{MBR}}_{t:t+H-1}
= \operatorname*{argmin}_{a^{(i)}_{t:t+H-1} \in \mathcal{A}} \; \mathcal{L}(a^{(i)}_{t:t+H-1}).
\label{eq:3}
\end{equation}
% We instantiate $u$ using agreement in predicted end-effector motion. For each sampled hypothesis $a^{(n)}_{t:t+H-1}$, we accumulate translational and rotational deltas to obtain a predicted trajectory, represented by a feature vector $\phi(a^{(n)}_{t:t+H-1}) \in \mathbb{R}^{6H}$ containing position $(x,y,z)$ and orientation $(u,v,w)$ at each step. We define the utility as negative distance: $u(y,h) = -d(\phi(y), \phi(h))$. Under this choice, MBR selects the consensus trajectory:

We instantiate $d$ using distance in predicted end-effector motion. For each sampled hypothesis $a^{(i)}_{t:t+H-1}$, we accumulate translational and rotational deltas to obtain a predicted trajectory, represented by a feature vector $\phi(a^{(i)}_{t:t+H-1}) \in \mathbb{R}^{6H}$ containing position $(x,y,z)$ and orientation $(u,v,w)$ at each step. MBR selects the consensus trajectory:
\begin{equation}
a^{\mathrm{MBR}}_{t:t+H-1}
= \operatorname*{argmin}_{a^{(i)}_{t:t+H-1} \in \mathcal{A}}
\frac{1}{N} \sum_{j=1}^{N}
d\!\left(\phi(a^{(i)}_{t:t+H-1}), \phi(a^{(j)}_{t:t+H-1})\right),
\label{eq:mbr}
\end{equation}
which minimizes the average pairwise distance over the $N \times N$ distance matrix, favoring trajectories in high-density regions of the policy output space.

\begin{algorithm}[t]
\caption{CycleVLA Inference}
\label{alg:cyclevla}
\footnotesize
\begin{algorithmic}[1]
\REQUIRE Task instruction $g$, VLA policy $\pi_\theta$, VLM $\mathcal{V}$, progress threshold $\tau_p$, action chunk size $H$, sample number $N$, max retries $R$, timeout $T_{\max}$
\ENSURE Episode success / failure

\vspace{0.3em}
\STATE Decompose $g$ into subtask list $G=(g_1,\ldots,g_K)$ by $\mathcal{V}$
\STATE $k \gets 1$; $t \gets 0$; $\texttt{phase} \gets \textsc{InProgress}$;\\ action queue $\mathcal{Q}\gets\emptyset$; $r_{1:K}\gets 0$

\vspace{0.3em}
\WHILE{$k \le K$ \textbf{and} $t < T_{\max}$}
    \STATE Observe $o_t$
    \IF{$\mathcal{Q}=\emptyset$}
        \STATE Sample chunk $a_{t:t+H-1}\sim\pi_\theta(\cdot\mid o_t,g_k)$ and push into $\mathcal{Q}$
    \ENDIF
    \STATE Pop $a_t$ from $\mathcal{Q}$ and parse $a_t=(\tilde a_t, s_t, p_t)$ with $\tilde a_t\in\mathbb{R}^7$
    \STATE Execute $\tilde a_t$; observe $o_{t+1}$; $t \gets t + 1$
    \IF{episode succeeds}
        \STATE \textbf{return} success
    \ENDIF
    
    \vspace{0.3em}
    \IF{$\texttt{phase}=\textsc{InProgress}$ \textbf{and} $\textsc{Confirm}(p_t\ge\tau_p)$}
        \STATE $(j,\texttt{dec}) \gets \mathcal{V}(o_t,g_k,G)$, $\texttt{dec}\in\{\texttt{transit},\texttt{backtrack}\}$
         \IF{$\texttt{dec}=\texttt{backtrack}$ \textbf{and} $r_j < R$}
            \STATE $r_j \gets r_j + 1$
            \STATE Restore robot to start of $g_j$ via reverse execution; $\mathcal{Q}\gets\emptyset$
            \STATE Observe $o_t'$ after restoration
            \STATE Sample $\mathcal{A}=\{a^{(i)}_{t:t+H-1}\}_{i=1}^{N}$ with $a^{(i)}_{t:t+H-1}\sim\pi_\theta(\cdot\mid o_t',g_j)$
            \STATE Select $a^{\mathrm{MBR}}_{t:t+H-1}$ via Eq.~\eqref{eq:mbr} and set $\mathcal{Q}\gets a^{\mathrm{MBR}}_{t:t+H-1}$
            \STATE $k\gets j$ 
        \ELSE
            \STATE $\texttt{phase}\gets\textsc{Complete}$
        \ENDIF

    \vspace{0.3em}
    \ELSIF{$\texttt{phase}=\textsc{Complete}$ \textbf{and} $\textsc{Confirm}(s_t=1)$}
        \STATE $k\gets k+1$; $\texttt{phase}\gets\textsc{InProgress}$; $\mathcal{Q}\gets\emptyset$
    \ENDIF
\ENDWHILE

\vspace{0.3em}
\STATE \textbf{return} failure
\end{algorithmic}
\end{algorithm}

\vspace{0.5em}
\noindent\textbf{Overall Inference Procedure.} We summarize the complete CycleVLA inference procedure in Alg.~\ref{alg:cyclevla}. The algorithm alternates between two phases per subtask: \textsc{InProgress}, where the robot executes until predicted progress triggers the VLM check, and \textsc{Complete}, where execution continues until the stop signal confirms subtask termination.

%------------------------------------------------------------------
\section{Simulation Experiments and Analysis}
We conduct experiments on simulation benchmarks to answer the following questions: 1) To what extent does our proactive self-correcting VLA improve task success rates, and how many failed trajectories can it recover compared to a monolithic VLA, even with under-trained VLAs? 2) How much does test-time scaling via MBR decoding contribute to performance gains in VLA inference? 3) What is the impact of each component on overall effectiveness?

% Next paper verison todo: 1) Real robot experiment, 2) Add pi-0.5 backbone, 3) Add more dataset (e.g., Calvin, RL-Bench, BridgeV2)

% Additional Analysis (Appendix)
% 1) Quality of decomposed dataset: human study. Qualitative results.
% 2) Effects of finetuning on decomposed subtask dataset with stop signal as a data augementation.
% 3) RL-based baseline.
% 4) During inference (real-robot), intentionally change the location of the object to create a failure. The VLA has to backtrack and retry.

\begin{table*}[t]
\setlength{\abovecaptionskip}{0.1cm}
    \footnotesize
    \caption{\textbf{LIBERO-Plus task performance} (success rates~$\uparrow$) under seven perturbation dimensions. Dark and light green cells denote the best and second-best results per column. Numbers in parentheses show the average improvement over the corresponding VLA backbone. $\dagger$\,OpenVLA-OFT-Diff is trained by us as a single policy across all LIBERO suites on third-person and wrist camera images and proprio, since the original OpenVLA-OFT~\cite{kim2025fine} does not report its diffusion variant in this setting. Please see Appendix~\ref{sec:Additional Experiments and Details} for more details on this baseline.}
    \begin{adjustbox}{width=0.85\linewidth,center}
    \centering
    \begin{tabular}{lcccccccc}
    \toprule
    \textbf{Method} & \textbf{Camera} & \textbf{Robot} & \textbf{Language} & \textbf{Light} & \textbf{Background} & \textbf{Noise} & \textbf{Layout} & \textbf{Average} \\ \noalign{\vskip 0.3ex}
    \hline \noalign{\vskip 1.0ex}
    OpenVLA\cite{KimPKXB0RFSVKBT24} & 0.8 & 3.5 & 23.0 & 8.1 & 34.8 & 15.2 & 28.5 & 16.3 \\
    \noalign{\vskip 0.5ex}
    UniVLA\cite{wang2025unifiedvisionlanguageactionmodel} & 1.8 & 46.2 & 69.6 & 69.0 & 81.0 & 21.2 & 31.9 & 45.8 \\
    \noalign{\vskip 0.5ex}
    OpenVLA-OFT-Diff\,$^\dagger$ & 40.3 & 40.0 & 79.2 & 77.9 & 79.6 & 68.1 & 71.6 & 65.2 \\
    \noalign{\vskip 0.5ex}
    % OpenVLA-OFT & 56.4 & 31.9 & 79.5 & 88.7 & 93.3 & 75.8 & 74.2 & 71.4 \\
    % \noalign{\vskip 0.5ex}
    $\pi_{0}$\cite{pi_levine} & 13.8 & 6.0 & 58.8 & 85.0 & 81.4 & \sbest{79.0} & 68.9 & 56.1 \\
    \noalign{\vskip 0.5ex}
    $\pi_{0}$-FAST\cite{pertsch2025fastefficientactiontokenization} & \sbest{65.1} & 21.6 & 61.0 & 73.2 & 73.2 & 74.4 & 68.8 & 62.5 \\
    \noalign{\vskip 0.5ex}
    WorldVLA\cite{cen2025worldvlaautoregressiveactionworld} & 0.1 & 27.9 & 41.6 & 43.7 & 17.1 & 10.9 & 38.0 & 25.6 \\
    \noalign{\vskip 0.5ex}
    $\pi_{0.5}$\cite{pi0_5_pi} & 53.0 & \sbest{50.3} & 65.7 & 83.1 & 77.3 & 53.2 & 72.7 & 65.0 \\
    \noalign{\vskip 0.5ex}
    \hline \noalign{\vskip 1.0ex}
    \textbf{CycleVLA} (OpenVLA-OFT-Diff) & 46.7 & 41.7 & \sbest{91.1} & \sbest{85.8} & \sbest{84.6} & 77.2 & \sbest{81.2} & \sbest{72.6}\,(+7.4) \\
    \noalign{\vskip 0.5ex}
    \textbf{CycleVLA} ($\pi_{0.5}$) & \best{74.2} & \best{72.9} & \best{96.7} & \best{96.5} & \best{91.3} & \best{86.1} & \best{84.5} & \best{86.0}\,(+21.0) \\
    \bottomrule
    \end{tabular}
    \label{tab:liberoplus}
\end{adjustbox}
\vspace{-0.5em}
\end{table*}

\subsection{Implementation Details}

\noindent\textbf{Subtask Decomposition.} We use GPT-4.1~\cite{GPT-4-1_openai} (temperature 0.2) to propose subtasks and infer their timestamp boundaries. The last action step of each subtask is oversampled by a factor of 8. Please see Appendix~\ref{sec:Details and Evaluation of Subtask Decomposition} for more details.

% To generate scene descriptions, we use Prismatic-7B VLM~\cite{Karamcheti0BLKS24}.

\vspace{0.5em}
\noindent\textbf{Training and Inference.} We test our approach using two VLA backbones: $\pi_{0.5}$~\cite{pi0_5_pi} and OpenVLA-OFT~\cite{kim2025fine} with a diffusion-based action head (OpenVLA-OFT-Diff). Stochastic sampling is achieved by varying random seeds, which changes the noise sampling in the diffusion/flow matching process. Models are trained on 4 or 8 NVIDIA A100 GPUs (40GB VRAM) and evaluated on 1 NVIDIA A10 GPU (24GB VRAM) unless otherwise specified. We use GPT-5.2~\cite{GPT-5-2_openai} (temperature 1.0) as the VLM-based failure predictor and planner, queried when subtask progress reaches $\tau_p = 0.9$. For MBR decoding, we use an $L_2$ distance metric ($d$) with $N=8$ sampled hypotheses. Unless otherwise specified, all analyses beyond the main LIBERO~\cite{LiuZGFLZS23} and LIBERO-Plus~\cite{fei2026liberoplus} comparisons use the OpenVLA-OFT-Diff-based CycleVLA on LIBERO. Please see Appendices~\ref{sec:Implementation Details of MBR Decoding} and~\ref{sec:Additional Experiments and Details} for more details.

\subsection{Evaluation on Simulation Benchmarks}\label{sec:Evaluation on Simulation Benchmarks}

\noindent\textbf{Setup.} We evaluate on LIBERO~\cite{LiuZGFLZS23} (four task suites: Spatial, Object, Goal, and Long) and LIBERO-Plus~\cite{fei2026liberoplus}, which introduces seven dimensions of perturbation to probe CycleVLA's generalization and robustness under unseen conditions. For LIBERO, we follow OpenVLA~\cite{KimPKXB0RFSVKBT24} for data preprocessing and evaluation. Each suite contains 10 tasks, evaluated over 50 rollouts with 3 random seeds. For finetuning, we follow~\cite{ZhengLH0DKHY25, huang2025thinkact, fpc_yang} and train a single model jointly on data from all four suites, rather than training separate models per suite as in~\cite{KimPKXB0RFSVKBT24, kim2025fine, li2025cogvla}. Joint training increases task and scene diversity and thus presents a more challenging learning setting. We report results alongside baseline numbers regardless of the finetuning paradigm. For LIBERO-Plus, we evaluate the same LIBERO-trained models directly on the perturbed environments without additional finetuning, isolating generalization from perturbation-specific adaptation. Evaluation follows LIBERO-Plus protocol.

\begin{table}[t]
    \vspace{0.6em}
    \setlength{\abovecaptionskip}{0.1cm}
    \footnotesize
    \caption{\textbf{LIBERO task performance} (success rates~$\uparrow$). All policies compared here are trained via supervised finetuning on demonstrations without RL-based post-training. Color and parenthesis conventions and the definition of $\dagger$ follow Table~\ref{tab:liberoplus}.}
    % \vspace{-0.5em}
    \begin{adjustbox}{width=1.0\linewidth,center}
    \centering
    \begin{tabular}{lccccc}
    \toprule
    \textbf{Method} & \textbf{Spatial} & \textbf{Object} & \textbf{Goal} & \textbf{Long} & \textbf{Average} \\ \noalign{\vskip 0.3ex}
    \hline \noalign{\vskip 1.0ex}
    Diffusion Policy~\cite{ChiFDXCBS23} & 78.3 & 82.5 & 68.3 & 50.5 & 72.4 \\
    \noalign{\vskip 0.5ex}
    Octo-Base~\cite{GhoshWPBMDHK0LT24} & 78.9 & 85.7 & 84.6 & 51.1 & 75.1 \\
    \noalign{\vskip 0.5ex}
    OpenVLA~\cite{KimPKXB0RFSVKBT24} & 84.7 & 88.4 & 79.2 & 53.7 & 76.5 \\
    \noalign{\vskip 0.5ex}
    TraceVLA~\cite{ZhengLH0DKHY25} & 84.6 & 85.2 & 75.1 & 54.1 & 74.8 \\
    \noalign{\vskip 0.5ex}
    SpatialVLA~\cite{qu2025spatialvla} & 88.2 & 89.9 & 78.6 & 55.5 & 78.1 \\
    \noalign{\vskip 0.5ex}
    CoT-VLA~\cite{ZhaoLKFZWLMHFHL25} & 87.5 & 91.6 & 87.6 & 69.0 & 81.1 \\
    \noalign{\vskip 0.5ex}
    ThinkAct~\cite{huang2025thinkact} & 88.3 & 91.4 & 87.1 & 70.9 & 84.4 \\
    \noalign{\vskip 0.5ex}
    GR00T N1~\cite{gr00t_fan} & 94.4 & 97.6 & 93.0 & 90.6 & 93.9 \\
    \noalign{\vskip 0.5ex}
    OpenVLA-OFT-Diff\,$^\dagger$ & 91.7 & 92.2 & 90.9 & 88.4 & 90.8 \\
    \noalign{\vskip 0.5ex}
    % OpenVLA-OFT~\cite{kim2025fine} & 96.9 & 98.1 & 95.6 & 91.1 & 95.4 \\
    % \noalign{\vskip 0.5ex}
    $\pi_{0.5}$~\cite{pi0_5_pi} & \sbest{98.8} & \sbest{98.2} & \best{98.0} & 92.4 & \sbest{96.9} \\
    \noalign{\vskip 0.5ex}
    \hline \noalign{\vskip 1.0ex}
    \textbf{CycleVLA} (OpenVLA-OFT-Diff) & 97.6 & 98.1 & 91.7 & \sbest{93.6} & 95.3\,(+4.5) \\
    \noalign{\vskip 0.5ex}
    \textbf{CycleVLA} ($\pi_{0.5}$) & \best{99.6} & \best{99.3} & \sbest{97.5} & \best{97.6} & \best{98.5}\,(+1.6) \\
    \bottomrule
    \end{tabular}
    \label{tab:libero_success}
\end{adjustbox}
\vspace{-0.5em}
\end{table}

\begin{figure*}[t]
  \centering
  \footnotesize
  \setlength{\abovecaptionskip}{0.1cm}
  \includegraphics[width=1\linewidth]{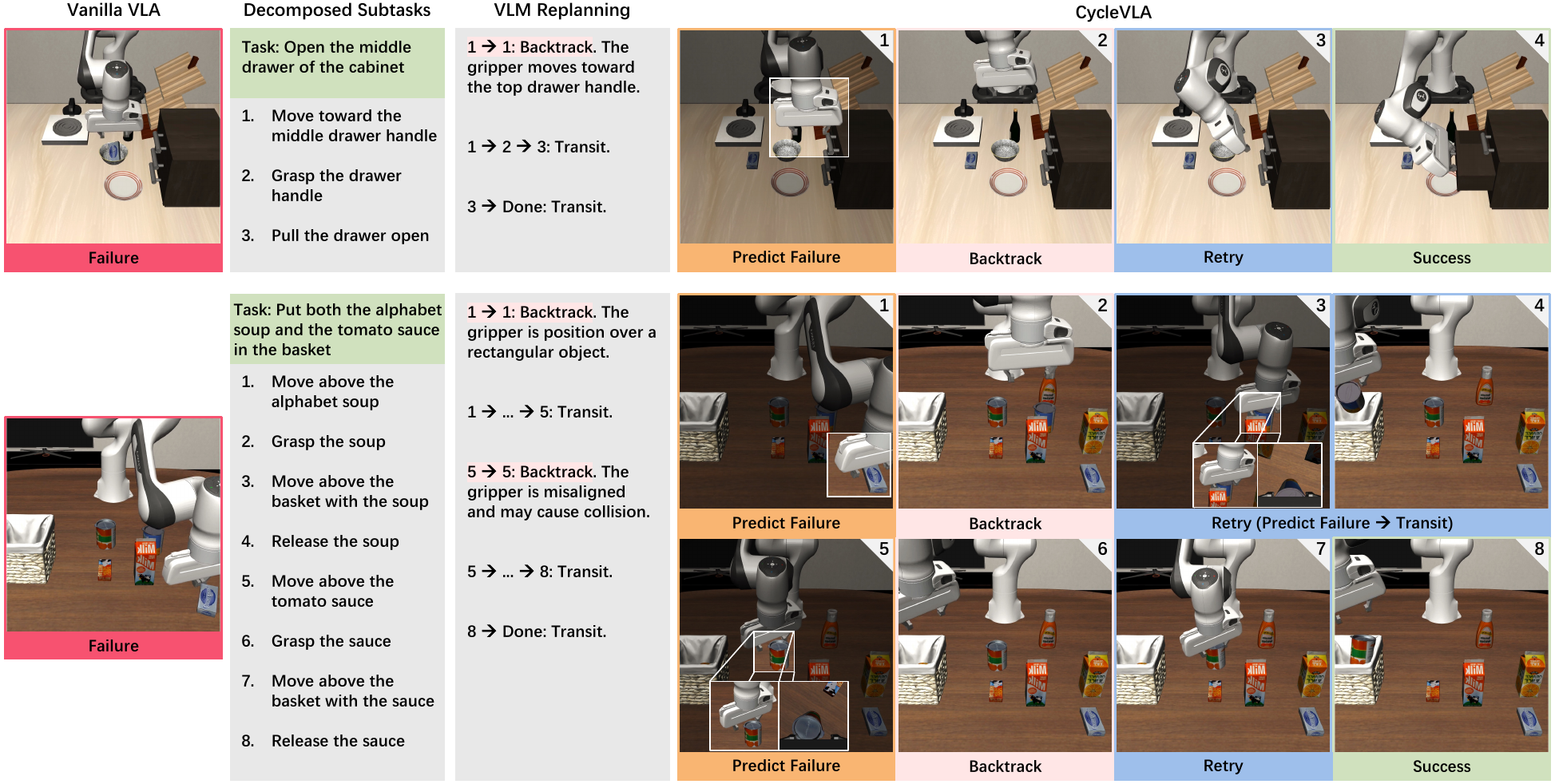}
  \vspace{-3mm} \caption{\textbf{Qualitative examples of CycleVLA in simulation.} CycleVLA performs multiple cycles of failure prediction, backtracking, and retry within a single long-horizon task, correcting errors across subtasks and achieving successful completion. More examples can be found in Appendix~\ref{sec:Additional Experiments and Details}.}
  \label{fig:demo}
  \vspace{-1.0em} 
\end{figure*}

\begin{table*}[b]
\setlength{\abovecaptionskip}{0.1cm}
    \footnotesize
    \caption{\textbf{Recovery performance on under-trained VLAs.} Success rates~$\uparrow$ on LIBERO with and without failure correction (FC).}
    \begin{adjustbox}{width=1.0\linewidth,center}
    \centering
    \begin{tabular}{lcccccccccc}
    \toprule
     \multirow{2}{*}[-0.5ex]{\textbf{Model}} & \multicolumn{2}{c}{\textbf{Spatial}} & \multicolumn{2}{c}{\textbf{Object}} & \multicolumn{2}{c}{\textbf{Goal}} & \multicolumn{2}{c}{\textbf{Long}} & \multicolumn{2}{c}{\textbf{Average}} \\
     \cmidrule(lr){2-3} \cmidrule(lr){4-5} \cmidrule(lr){6-7} \cmidrule(lr){8-9} \cmidrule(lr){10-11}
     & w/o FC & w/ FC & w/o FC & w/ FC & w/o FC & w/ FC & w/o FC & w/ FC & w/o FC & w/ FC \\ \noalign{\vskip 0.3ex}
    \hline \noalign{\vskip 1.0ex}
    CycleVLA-200K & 84.7 & 89.1 {\textcolor{improvementcolor}{(+4.4)}} & 82.0 & 90.2 {\textcolor{improvementcolor}{(+8.2)}} & 63.6 & 74.2 {\textcolor{improvementcolor}{(+10.6)}} & 62.3 & 66.5 {\textcolor{improvementcolor}{(+4.2)}} & 73.2 & 80.0 {\textcolor{improvementcolor}{(+6.8)}} \\
    \noalign{\vskip 0.5ex}
    CycleVLA-350K & 88.4 & 94.8 {\textcolor{improvementcolor}{(+6.4)}} & 93.8 & 97.5 {\textcolor{improvementcolor}{(+3.7)}} & 72.6 & 79.4 {\textcolor{improvementcolor}{(+6.8)}} & 77.9 & 85.3 {\textcolor{improvementcolor}{(+7.4)}} & 83.2 & 89.2 {\textcolor{improvementcolor}{(+6.0)}} \\
    \noalign{\vskip 0.5ex}
    CycleVLA-500K & 91.1 & 97.6 {\textcolor{improvementcolor}{(+6.5)}} & 95.3 & 98.1 {\textcolor{improvementcolor}{(+2.8)}} & 86.5 & 91.7 {\textcolor{improvementcolor}{(+5.2)}} & 84.4 & 93.6 {\textcolor{improvementcolor}{(+9.2)}} & 89.3 & 95.3 {\textcolor{improvementcolor}{(+6.0)}} \\
    \bottomrule
    \end{tabular}
    \label{tab:libero_recovery}
\end{adjustbox}
% \vspace{-0.5em}
\end{table*}

\begin{table*}[b]
\setlength{\abovecaptionskip}{0.1cm}
    \footnotesize
    \caption{\textbf{Effect of number of hypotheses on MBR performance.} Estimated success probability ($P_{\text{succ}}$~$\uparrow$). w/o FC denotes the average success rate of the base VLA on LIBERO without failure correction.}
    \begin{adjustbox}{width=1.0\linewidth,center}
    \centering
    \begin{tabular}{lccccccccccc}
    \toprule
    \multirow{2}{*}[-0.5ex]{\textbf{Model}} &
    \multirow{2}{*}[-0.5ex]{\textbf{w/o FC}} &
    \multicolumn{2}{c}{\textbf{4}} &
    \multicolumn{2}{c}{\textbf{8}} &
    \multicolumn{2}{c}{\textbf{16}} &
    \multicolumn{2}{c}{\textbf{32}} &
    \multicolumn{2}{c}{\textbf{64}} \\
    \cmidrule(lr){3-4} \cmidrule(lr){5-6} \cmidrule(lr){7-8} \cmidrule(lr){9-10} \cmidrule(lr){11-12}
    & & random & MBR & random & MBR & random & MBR & random & MBR & random & MBR \\ 
    \noalign{\vskip 0.3ex}
    \hline
    \noalign{\vskip 1.0ex}
    CycleVLA-200K & 73.2 & 72.9 & 78.2 {\textcolor{improvementcolor}{(+5.3)}} & 71.7 & 78.5 {\textcolor{improvementcolor}{(+6.8)}} & 71.9 & 81.2 {\textcolor{improvementcolor}{(+9.3)}} & 72.2 & 79.7 {\textcolor{improvementcolor}{(+7.5)}} & 72.3 & 79.7 {\textcolor{improvementcolor}{(+7.4)}} \\
    \noalign{\vskip 0.5ex}
    CycleVLA-350K & 83.2 & 80.5 & 88.0 {\textcolor{improvementcolor}{(+7.5)}} & 80.3 & 90.2 {\textcolor{improvementcolor}{(+9.9)}} & 80.2 & 91.3 {\textcolor{improvementcolor}{(+11.1)}} & 80.4 & 92.3 {\textcolor{improvementcolor}{(+11.9)}} & 80.7 & 92.2 {\textcolor{improvementcolor}{(+11.5)}} \\
    \noalign{\vskip 0.5ex}
    CycleVLA-500K & 89.3 & 90.8 & 94.1 {\textcolor{improvementcolor}{(+3.3)}} & 90.2 & 95.5 {\textcolor{improvementcolor}{(+5.3)}} & 91.1 & 95.7 {\textcolor{improvementcolor}{(+4.6)}} & 90.7 & 95.7 {\textcolor{improvementcolor}{(+5.0)}} & 90.4 & 95.6 {\textcolor{improvementcolor}{(+5.2)}} \\
    \bottomrule
    \end{tabular}
    \label{tab:mbr_hypotheses}
\end{adjustbox}
\vspace{-0.5em}
\end{table*}

\vspace{0.5em}
\noindent\textbf{LIBERO Task Performance.} As shown in Table~\ref{tab:libero_success}, prior methods often achieve lower success rates on LIBERO-Long due to its challenging long-horizon nature. CycleVLA achieves notably higher success rates on this task suite by predicting failures at subtask boundaries and backtracking to retry. This self-correcting mechanism is particularly effective for long-horizon tasks, where errors tend to accumulate across subtasks. CycleVLA improves both backbones, with average gains of +4.5 over OpenVLA-OFT-Diff and +1.6 over $\pi_{0.5}$, the latter reaching the best average success rate (98.5) among all compared methods. OpenVLA-OFT-Diff benefits more, as $\pi_{0.5}$'s near-saturated performance leaves less room for correction. We further study how much CycleVLA improves less robust VLAs in the under-trained VLA analysis below.

\vspace{0.5em}
\noindent\textbf{LIBERO-Plus Task Performance.} From Table~\ref{tab:liberoplus}, CycleVLA improves robustness under unseen perturbations for both VLA backbones, with average gains of +7.4 over OpenVLA-OFT-Diff and +21.0 over $\pi_{0.5}$, considerably larger than those on clean LIBERO. We attribute this to the nature of perturbation-induced failures: perturbations rarely eliminate the policy's ability to act, but instead cause recoverable execution errors such as misgrasps or misplaced objects. A monolithic VLA commits to such errors, whereas CycleVLA detects them at subtask boundaries and retries with newly sampled, MBR-selected actions, converting near-misses into successes. Unlike on clean LIBERO, the stronger $\pi_{0.5}$ backbone gains more here: being more capable, it typically knows what to do under perturbations and merely fails occasionally in execution, producing exactly the recoverable errors CycleVLA is designed to correct. Conversely, the gains are more modest under camera and robot-state perturbations for the OpenVLA-OFT-Diff-based variant, likely because these shifts more directly affect the underlying policy’s perception and control. In such cases, failure correction can recover from some execution errors, but its benefit is naturally limited when the base policy does not complete the subtask reliably.

\vspace{0.5em}
\noindent\textbf{Effectiveness on Under-Trained VLAs.} In practice, due to computational constraints or the high difficulty of certain robotic tasks, some VLAs may be under-trained. We investigate whether CycleVLA remains effective across VLAs with varying capacities. We select two intermediate checkpoints (trained after 200K and 350K steps) alongside our final 500K checkpoint, and compare performance before and after applying failure correction. In the no-correction setting, the policy naturally progresses to the next subtask when a stop signal is triggered. Results in Table~\ref{tab:libero_recovery} show that CycleVLA provides consistent gains across all checkpoints. Notably, CycleVLA bridges the gap between model sizes: the 200K and 350K checkpoints with CycleVLA approach the performance of the 350K and 500K checkpoints without it.

\vspace{0.5em}
\noindent\textbf{Qualitative Results.} We show qualitative examples of CycleVLA in Fig.~\ref{fig:demo} on tasks with varying horizons.

\begin{table*}[t]
\setlength{\abovecaptionskip}{0.1cm}
    \footnotesize
    \caption{\textbf{Effect of distance metric on MBR performance.} Estimated success probability ($P_{\text{succ}}$~$\uparrow$). w/o FC denotes the average success rate of the base VLA on LIBERO without failure correction.}
    \begin{adjustbox}{width=0.85\linewidth,center}
    \centering
    \begin{tabular}{lccccccc}
    \toprule
    \textbf{Model} & \textbf{w/o FC} & \textbf{Random} & 
    \textbf{MBR-$L_1$} & 
    \textbf{MBR-$L_2$} & 
    \textbf{MBR-$L_\infty$} & 
    \textbf{MBR-$\cos$} &
    \textbf{MBR-$r$}\\
    \noalign{\vskip 0.3ex}
    \hline
    \noalign{\vskip 1.0ex}
    CycleVLA-200K & 73.2 & 71.7 & \textbf{78.7 {\textcolor{improvementcolor}{(+7.0)}}} & 78.5 {\textcolor{improvementcolor}{(+6.8)}} & 77.8 {\textcolor{improvementcolor}{(+6.1)}} & 74.1 {\textcolor{improvementcolor}{(+2.4)}} & 73.5 {\textcolor{improvementcolor}{(+1.8)}} \\
    CycleVLA-350K & 83.2 & 80.3 & 89.7 {\textcolor{improvementcolor}{(+9.4)}} & \textbf{90.2 {\textcolor{improvementcolor}{(+9.9)}}} & 88.2 {\textcolor{improvementcolor}{(+7.9)}} & 87.5 {\textcolor{improvementcolor}{(+7.2)}} & 86.9 {\textcolor{improvementcolor}{(+6.6)}} \\
    CycleVLA-500K & 89.3 & 90.2 & 94.7 {\textcolor{improvementcolor}{(+4.5)}} & \textbf{95.5 {\textcolor{improvementcolor}{(+5.3)}}} & 94.3 {\textcolor{improvementcolor}{(+4.1)}} & 94.4 {\textcolor{improvementcolor}{(+4.2)}} & 94.8 {\textcolor{improvementcolor}{(+4.6)}} \\
    \bottomrule
    \end{tabular}
    \label{tab:mbr_metrics}
\end{adjustbox}
% \vspace{-0.5em}
\end{table*}

\subsection{Analysis of MBR Decoding}\label{sec:Analysis of MBR Decoding}

\noindent\textbf{Setup.} We evaluate MBR decoding as a test-time scaling strategy for VLAs by measuring its ability to select successful action chunks from multiple stochastic hypotheses. For a given base VLA and MBR hyperparameter setting, results are averaged over 2 randomly selected tasks from each LIBERO task suite (8 tasks in total).

For each task, we evaluate on $E=200$ episodes. In each episode, we execute the VLA for $N$ stochastic trials, following the same execution protocol as the no-correction setting. We record the resulting trajectories and their success outcomes, and treat these $N$ executions as hypotheses. For each episode $e$, we define the set of decision steps (chunk boundaries) as $\mathcal{T}_e = \{0, H, 2H, \ldots\}$, where $H$ is the action chunk size. Each recorded trajectory is divided accordingly, yielding a temporally aligned set of chunk hypotheses $\{a^{(i)}_{t:t+H-1}\}_{i=1}^N$ at each step $t \in \mathcal{T}_e$.

At each step $t \in \mathcal{T}_e$, we compare two selection strategies over the same hypothesis set: 1) MBR decoding using Eq.~\eqref{eq:mbr}, and 2) random selection of a single hypothesis. Let $z^{(i)}_{e,t} \in \{0,1\}$ indicate whether hypothesis $i$ is successful at decision step $t$ in episode $e$ (inherited from the recorded rollout). For a selection method $m \in \{\textsc{Mbr}, \textsc{Random}\}$, let $\hat{\imath}^{(m)}_{e,t} \in \{1,\ldots,N\}$ denote the selected hypothesis index at step $t$. For each episode, we compute a score by averaging the success of the selected chunks over all decision steps. The final estimated success probability is then obtained by averaging over episodes:
\begin{equation}
P_{\text{succ}}^{(m)}
= \frac{1}{E}\sum_{e=1}^{E}
\left(
\frac{1}{|\mathcal{T}_e|}\sum_{t\in\mathcal{T}_e}
z^{(\hat{\imath}^{(m)}_{e,t})}_{e,t}
\right).
\label{eq:5}
\end{equation}
This metric estimates the probability of choosing a successful action chunk from $N$ sampled hypotheses, averaged across trajectory timestamps and repeated runs.

\begin{table}[t]
\vspace{0.6em}
\setlength{\abovecaptionskip}{0.1cm}
    \footnotesize
    \caption{\textbf{Ablation study on LIBERO.} Averaged success rate after removing individual components. LAO = last-action oversampling; UB = upper bound; LB = lower bound.}
    % \vspace{-0.5em}
    \begin{adjustbox}{width=0.85\linewidth,center}
    \centering
    \begin{tabular}{lc}
    \toprule
    \textbf{Method} & \textbf{Success Rate~$\uparrow$} \\
    \midrule
    CycleVLA (w/o MBR)         & 92.5 \\
    \noalign{\vskip 0.5ex}
    CycleVLA (alt. VLM)        & 92.8 \\
    \noalign{\vskip 0.5ex}
    CycleVLA (w/o stop + LAO)  & 91.1 \\
    \noalign{\vskip 0.5ex}
    CycleVLA (always-on MBR, UB)   & 96.9 \\
    \noalign{\vskip 0.5ex}
    CycleVLA (pred. failure cutoff, LB)   & 79.7 \\
    \noalign{\vskip 0.5ex}
    CycleVLA (main)            & 95.3 \\
    \bottomrule
    \end{tabular}
    \label{tab:libero_ablation}
\end{adjustbox}
\vspace{-0.5em}
\end{table}

\vspace{0.5em}
\noindent\textbf{Number of Hypotheses.} We study the effect of the number of hypotheses by varying $N \in \{4, 8, 16, 32, 64\}$, using an $L_2$ distance metric. Increasing $N$ generally improves MBR performance by better approximating expected risk (Eq.~\eqref{eq:1}), but with diminishing returns and higher computational cost due to increased sampling and pairwise distance evaluation. As shown in Table~\ref{tab:mbr_hypotheses}, most gains are achieved when increasing $N$ from 4 to 8, while improvements plateau beyond $N=16$, with similar performance for $N=32$ and $64$.

Without MBR decoding, random selection achieves a success probability similar to the base VLA because it samples action chunks directly from the model’s stochastic output distribution. Averaging performance across chunks, episodes, and tasks therefore reflects the marginal behavior of the underlying policy, without introducing any bias toward higher-quality candidates.

In contrast, MBR decoding consistently improves success probability across VLAs of varying capability, with larger gains observed for under-trained models (200K and 350K steps) compared to the fully trained 500K model. This trend aligns with observations in LLMs that weaker models benefit more from MBR-style selection~\cite{HeinemanD024}, as stronger models tend to produce more self-consistent candidates, reducing the marginal benefit of hypothesis selection.

\vspace{0.5em}
\noindent\textbf{Choice of Distance Metric.} We analyze the sensitivity of MBR decoding to the choice of distance metric $d$ used in the pairwise distance computation (Eq.~\eqref{eq:mbr}). We evaluate common metrics, including $L_1$, $L_2$, Chebyshev ($L_\infty$), cosine similarity ($\cos$), and correlation ($r$), with the number of hypotheses fixed to $N=8$. As shown in Table~\ref{tab:mbr_metrics}, $L_2$ consistently achieves the best performance across all VLAs, followed by $L_1$, while $\cos$ and $r$ yield the smallest improvements. We hypothesize this is because translational components are dense along the trajectory, whereas rotational components are sparse, with limited rotation at many timesteps. Distance-based metrics such as $L_1$ and $L_2$ therefore better capture magnitude differences in action sequences, while $\cos$ and $r$ emphasize directional agreement.

\begin{figure}[t]
  \centering
  \footnotesize
  \setlength{\abovecaptionskip}{0.1cm}
  \includegraphics[width=1\linewidth]{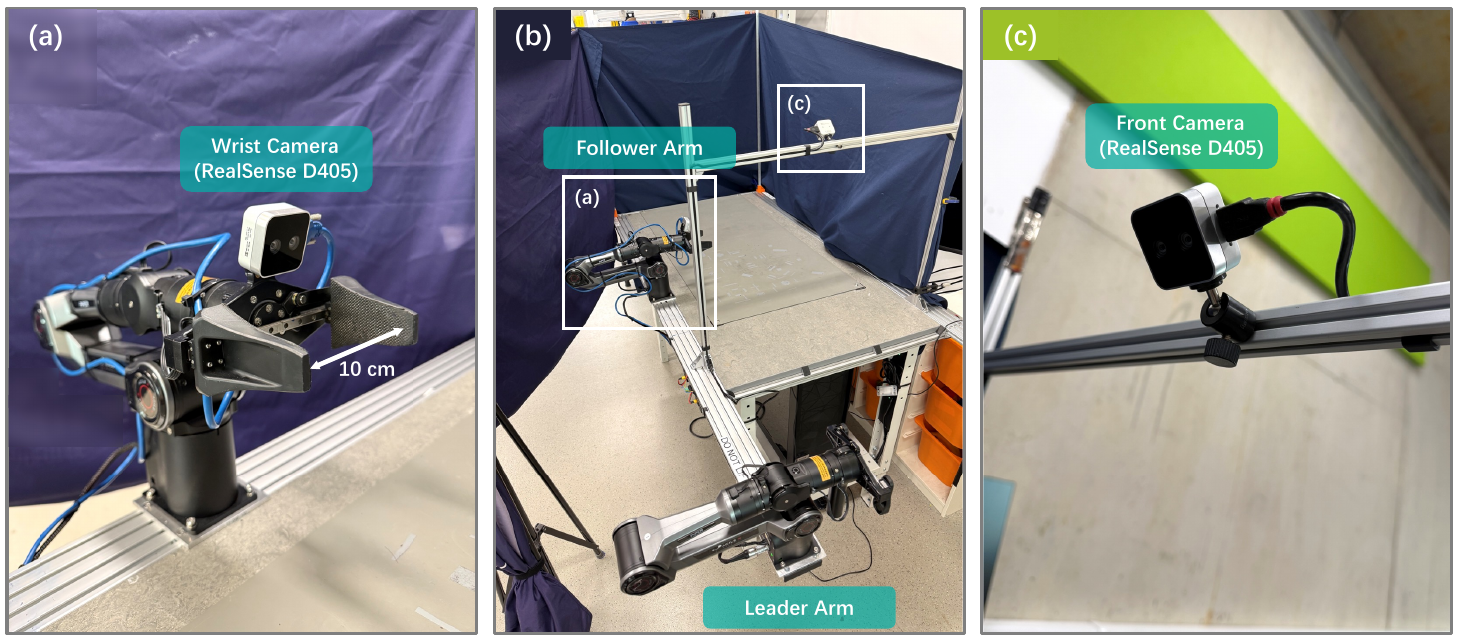}
  \vspace{-3mm} \caption{\textbf{System setup.} (a) Follower arm end-effector with a wrist-mounted RealSense D405 camera and a gripper with 10\,cm opening. (b) Workspace overview: the follower arm, and the leader arm operated by a human for teleoperated demonstration collection. (c)~Front-view camera (RealSense D405) mounted above the workspace.}
  \label{fig:setup}
  \vspace{-1.0em} 
\end{figure}

\subsection{Ablation Studies}\label{sec:Ablation Studies}
We evaluate the effectiveness of each component by removing or replacing it with an alternative design. \textbf{1) Removing MBR decoding:} After backtracking, MBR decoding is replaced with random hypothesis selection. \textbf{2) Changing VLM backbone:} We replace GPT-5.2~\cite{GPT-5-2_openai} with the smaller LLaMA-3.2-11B~\cite{llama3_Dubey} to assess robustness. \textbf{3) Removing stop signal and last-action oversampling:} We remove stop signal prediction and last-action oversampling during finetuning, reducing the action to $a_t = [\Delta x_t, \Delta y_t, \Delta z_t, \Delta u_t, \Delta v_t, \Delta w_t, \gamma_t, p_t]^\top \in \mathbb{R}^8$, and consider a subtask complete when $p_t \ge 0.95$ at inference. \textbf{4) Always-on MBR decoding (upper bound):} MBR decoding is applied at initial execution and every high-progress subtask transitions, without using the VLM to decide if to transit or backtrack. \textbf{5) Predicted failure cutoff (lower bound):} When the VLM predicts a potential failure, the episode is immediately terminated and counted as a failure. In principle, the resulting success rate should match that of the VLA without failure correction (Table~\ref{tab:libero_recovery}), and thus serves to evaluate the robustness of VLM failure prediction.

From Table~\ref{tab:libero_ablation}, removing MBR leads to a moderate drop in success rate, as retrying up to $R$ times (Alg.~\ref{alg:cyclevla}) already provides partial robustness. Using the generally less powerful LLaMA-3.2-11B reduces success rate, as we observe that it more frequently chooses to transit rather than backtrack, missing correction opportunities. Removing stop signal and last-action oversampling degrades performance due to spurious high-progress predictions causing premature termination. The always-on MBR variant achieves the highest success rate, but this comes at a steep efficiency trade-off, requiring $\sim$2.2$\times$ the inference time of our main method from MBR computation at every subtask transition; our VLM-gated design retains most of the gain at a fraction of the cost. Finally, terminating upon VLM-predicted failure causes a modest $\sim$10\% drop, illustrating LLM/VLM sycophancy~\cite{SharmaTKDABDHJK24, qiu2023efficient, qiu2023vfedsec, ma2023gradient}: when asked to predict failure, the VLM tends to confirm this assumption.

% From Table~\ref{tab:libero_ablation}, removing MBR leads to a moderate drop in success rate, as retrying up to $R$ times (Alg.~\ref{alg:cyclevla}) already provides partial robustness at the cost of runtime. Using LLaMA-3.2-11B reduces both success rate and runtime, as local inference is faster and we observe that LLaMA-3.2-11B more frequently chooses to transit rather than backtrack, leading to less action sampling and MBR computation. Removing stop signal and last-action oversampling reduces both metrics due to spurious high-progress predictions causing premature termination. The always-on MBR variant achieves higher success but incurs substantially higher runtime from MBR computation at every transition. Finally, terminating upon VLM-predicted failure causes a modest $\sim$10\% drop, illustrating LLM/VLM sycophancy~\cite{SharmaTKDABDHJK24, qiu2023efficient, qiu2023vfedsec, ma2023gradient}: when asked to predict failure, the VLM tends to confirm this assumption.

%------------------------------------------------------------------
\section{Real-World Experiments}
We validate CycleVLA on an AgileX PiPER arm, addressing two questions: 1) How does CycleVLA perform on a real robot, and can it detect and recover from errors arising from the VLA policy itself? 2) Under stress testing, where a human manually injects diverse error types, can CycleVLA still recover?

\subsection{Implementation Details}

\noindent\textbf{System Setup and Demonstration Collection.} We use an AgileX PiPER 6-DoF arm in a leader-follower configuration (Fig.~\ref{fig:setup}). Demonstrations are collected via teleoperation, where a human operator moves the leader arm and the follower arm mirrors its motion. During policy execution, the leader-follower mode is disabled and only the follower arm is actuated. Visual observations are captured by two Intel RealSense D405 cameras: a front camera mounted above the workspace and a wrist camera on the follower end-effector. See Appendix~\ref{sec:Additional Details of Real System} for further details.

\begin{figure}[t]
  \centering
  \footnotesize
  \setlength{\abovecaptionskip}{0.1cm}
  \includegraphics[width=1\linewidth]{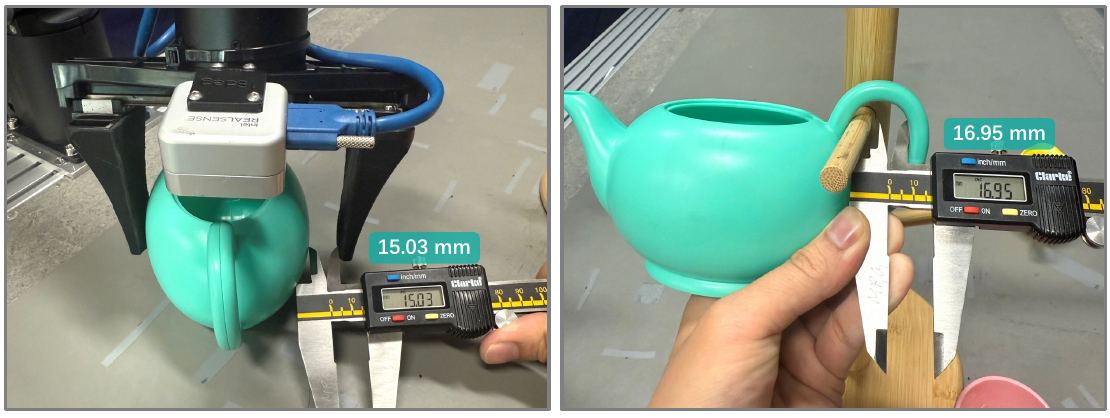}
  \vspace{-3mm} \caption{\textbf{Tight tolerances in the teapot-hanging task.} Left: with the teapot body grasped, the remaining gripper clearance is only 1.5\,cm. Right: with the peg inserted through the handle, the remaining clearance is only 1.7\,cm, requiring accurate positioning when hanging.}
  \label{fig:precise}
  \vspace{-1.0em} 
\end{figure}

\vspace{0.5em}
\noindent\textbf{Training and Inference.} For the real-robot policy, we use $\pi_{0.5}$~\cite{pi0_5_pi}, trained on 8 NVIDIA A100 GPUs (40GB VRAM) and evaluated on a robot-local NVIDIA RTX 3090 GPU (24GB VRAM). Inference parameters follow the simulation setup. See Appendices~\ref{sec:Additional Experiments and Details} and~\ref{sec:Additional Details of Real System} for more details.

\begin{figure*}[t]
  \centering
  \footnotesize
  \setlength{\abovecaptionskip}{0.1cm}
  \includegraphics[width=1\linewidth]{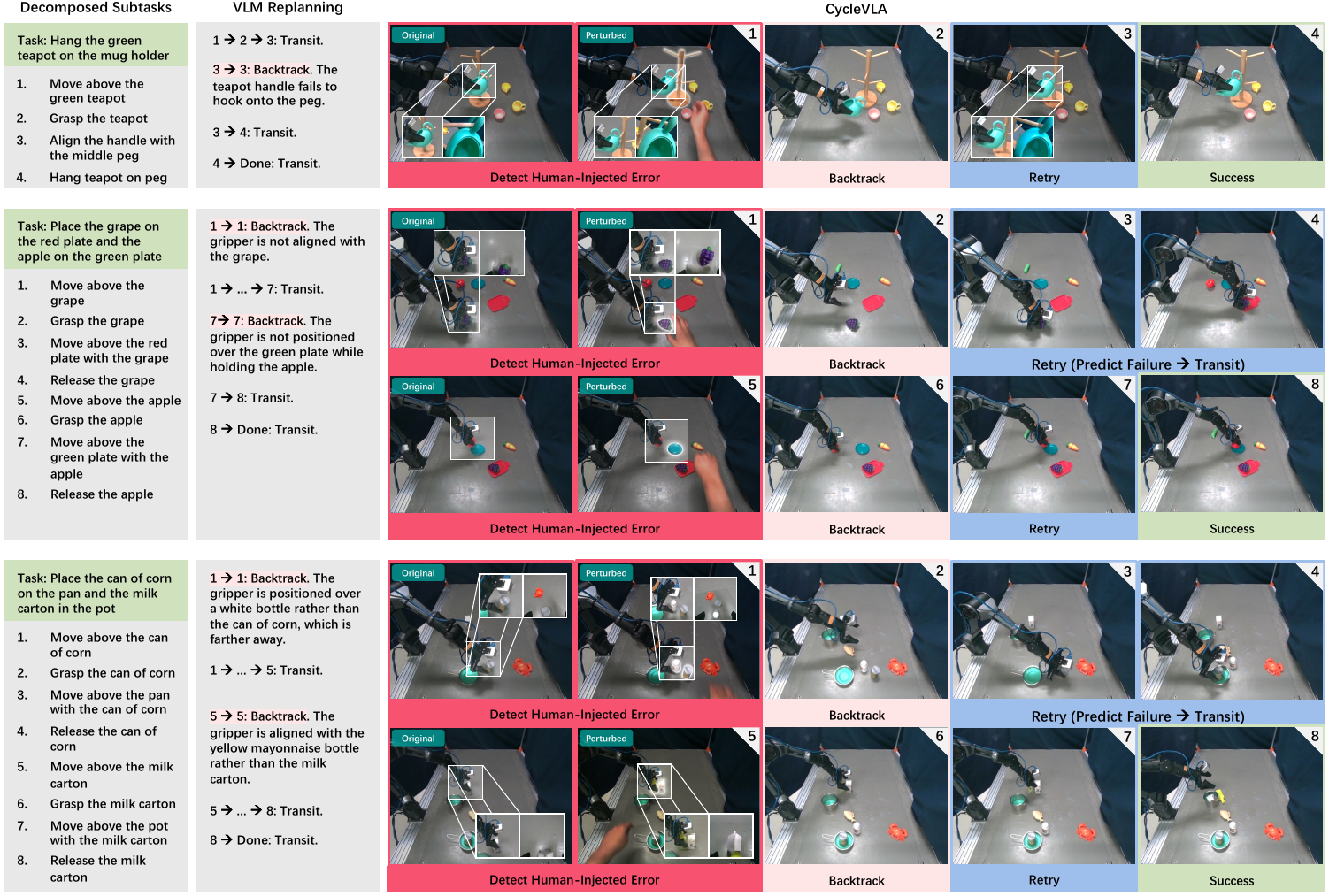}
    \vspace{-3mm} \caption{\textbf{Qualitative results of CycleVLA on a real robot.} Errors are injected via manual perturbations during execution. First row: a human perturbs the position and orientation of the mug holder and unhooks the teapot. Second row: a human displaces the grape and the green plate just as the robot is poised to grasp the grape and release the apple. Third row: a human relocates the corn can and the milk carton, placing a similar-looking pepper bottle and a mayonnaise bottle at their original locations, respectively.}
  \label{fig:demo_real}
  % \vspace{-1.0em} 
\end{figure*}

\subsection{Evaluation Protocols: Natural Failures and Injected Errors}
\noindent\textbf{Tasks and Setup.} We evaluate on three tasks:
\begin{itemize}
\renewcommand\labelitemi{\scalebox{0.75}{$\bullet$}}
\item \textit{Hang the green teapot on the mug holder.} A precision task where the teapot handle must pass over the wooden peg and the teapot body is only slightly narrower than the maximum gripper opening, requiring an accurate grasp position and orientation (Fig.~\ref{fig:precise}).
\item \textit{Place the grape on the red plate and the apple on the green plate.} A long-horizon task where errors are likely to accumulate across subtasks.
\item \textit{Place the can of corn on the pan and the milk carton in the pot.} A long-horizon task involving reflective cookware surfaces, which challenge visual perception.
\end{itemize}
For each task, we collect 100 teleoperated demonstrations with object positions randomized within $\pm$30\,cm and orientations within $\pm$60$^\circ$; subtask timestamps are recorded during teleoperation. Each setting is evaluated over 15 trials with randomized object positions and rotations. We evaluate CycleVLA under two settings of increasing difficulty: standard execution with natural failures, and stress testing with human-injected errors.

\vspace{0.5em}
\noindent\textbf{Standard Execution: Natural Failures.} In this evaluation setting, the robot executes each task autonomously, and failures arise only from the policy's own execution. We compare CycleVLA against a baseline without failure correction, where the trained VLA transitions to the next subtask whenever it predicts a 
stop signal.

\vspace{0.5em}
\noindent\textbf{Stress Testing: Human-Injected Errors.} In this more challenging setting, a human manually injects an error into one subtask per trial. Since the closed-loop policy naturally absorbs early perturbations (e.g., the robot simply re-approaches a moved object), we inject errors near subtask completion, just before the progress threshold triggers the VLM query, so that each perturbation constitutes a genuine failure requiring correction. This setting directly measures how much CycleVLA can rescue: the VLM must detect the induced failure and backtrack to an appropriate earlier subtask rather than transitioning forward. For each task, we inject three progressively more challenging error types:
\begin{itemize}
\renewcommand\labelitemi{\scalebox{0.75}{$\bullet$}}
\item \textbf{Distractor Injection.} The target object remains unchanged, but we place a visually similar distractor nearby to test whether the robot distinguishes the correct target. For example, we add a pink teapot and a yellow mug as the robot approaches the green teapot.
\item \textbf{Target Displacement.} We perturb the position and orientation of the placement target during interaction and evaluate whether the robot detects the mismatch and backtracks. For instance, while the robot holds the teapot with the handle around the peg, we move and rotate the mug holder so the peg no longer passes through the handle; similarly, we relocate the plate as the robot is about to release the apple onto it.

\begin{figure}[t]
  \centering
  \footnotesize
  \setlength{\abovecaptionskip}{0.1cm}
  \includegraphics[width=1\linewidth]{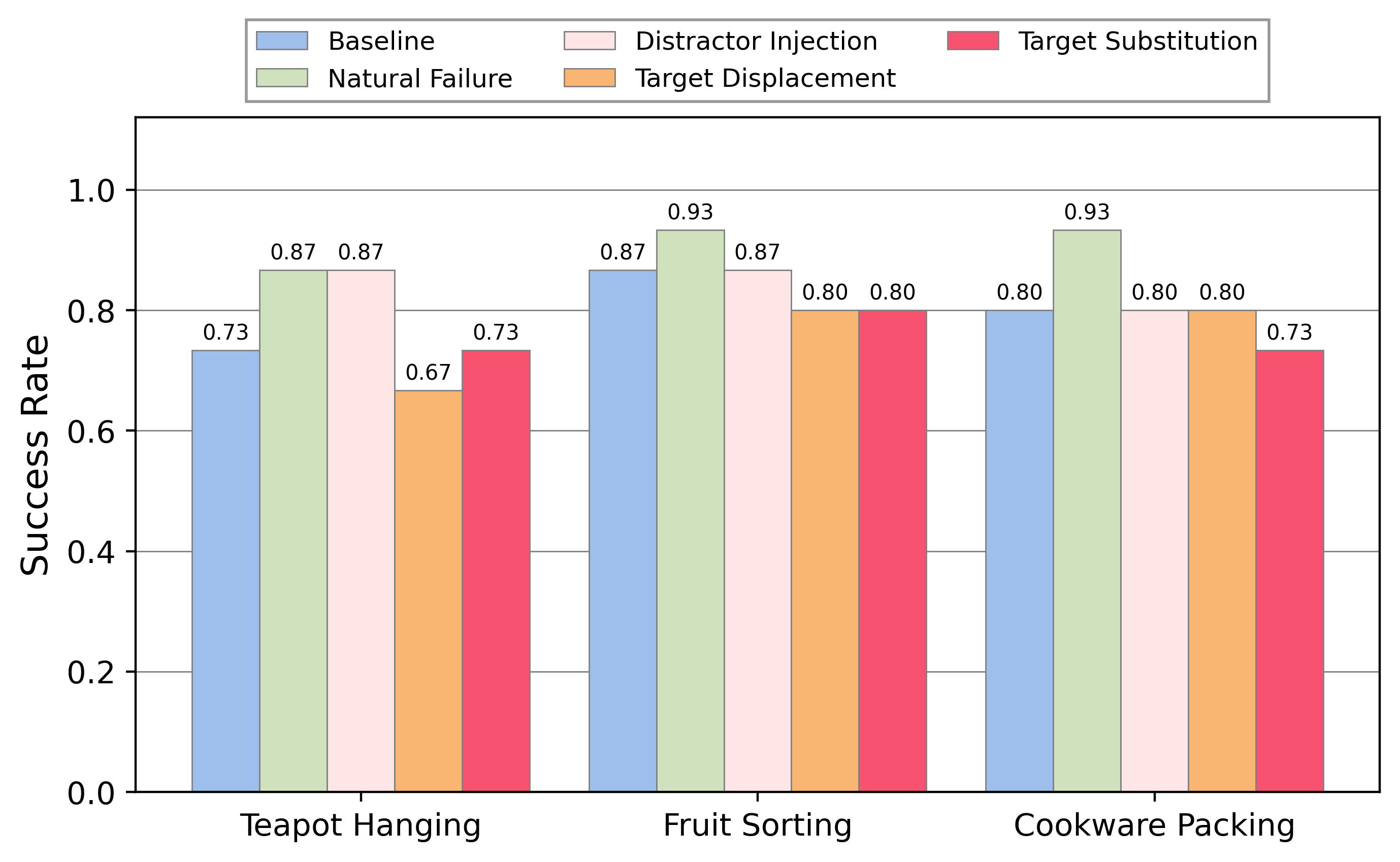}
    \vspace{-3mm} \caption{\textbf{Real-robot evaluation} (success rates~$\uparrow$). The baseline uses the same policy without failure correction. CycleVLA is evaluated in the natural-failure setting and under three injected perturbations: distractor injection, target displacement, and target substitution. CycleVLA improves success rates over the baseline in the natural-failure setting and maintains robust performance across all perturbations.}
  \label{fig:real_robot_result}
  \vspace{-1.0em} 
\end{figure}

\item \textbf{Target Substitution.} We replace the target object with an unrelated one and move the original target elsewhere. This error compounds the previous two, as a new object appears at the expected location while the true target is displaced. For example, as the robot is poised to grasp the grape, we swap in an eggplant under the gripper and relocate the grape.
\end{itemize}

\vspace{0.5em}
\noindent\textbf{Results and Analysis.} Results are shown in Fig.~\ref{fig:real_robot_result}. CycleVLA in the natural-failure setting consistently outperforms the baseline across all three tasks (0.87 vs.\ 0.73, 0.93 vs.\ 0.87, and 0.93 vs.\ 0.80), confirming that failure prediction and retry transfer to real-world execution. The teapot hanging task yields slightly lower success rates than the other tasks, as hanging the teapot onto the mug holder requires precise alignment. Under stress testing, since every trial contains an injected error, the success rate directly reflects the fraction of failures corrected: CycleVLA corrects $\sim$80\% of injected failures on average across all perturbation types and tasks. Across perturbation types, success rates decrease progressively from distractor injection to target displacement to target substitution, reflecting their increasing difficulty. Under target displacement, teapot hanging drops to 0.67 since the perturbation also relocates the mug holder, forcing the robot to retry the precise hanging motion at a new position. Under target substitution, all tasks use visually similar distractors (a pink pot for the teapot, an eggplant for the grape, and a red sausage for the apple), yet cookware packing scores lower than fruit sorting: the policy struggles to distinguish the corn can from the substituted white pepper bottle, possibly aggravated by reflections from the shiny pot wires near the target region. Qualitative examples of these correction behaviors are illustrated in Fig.~\ref{fig:demo_real}.

%------------------------------------------------------------------
\section{Conclusion and Discussion}
We introduce \textbf{CycleVLA}, a system that equips VLAs with proactive self-correction, the capability to anticipate incipient failures and recover before execution collapses. Experiments on simulation benchmarks and a real robot show that CycleVLA improves success rates by correcting execution failures across VLAs of varying capability, and remains robust under manually injected perturbations. We see our work as an early investigation into granting VLAs such abilities and exploring test-time scaling for VLAs. CycleVLA relies on an external VLM for failure prediction and replanning, and an explicit backtracking mechanism. Future work may investigate integrating these capabilities directly into VLAs to enable end-to-end failure reasoning and learned recovery behaviors, as well as tailored test-time scaling strategies for VLAs.

\vspace{0.5em}
\noindent\textbf{Limitations.} Our backtracking mechanism assumes reversible robot state transitions, which may not hold in highly dynamic or irreversible environments. Test-time scaling via MBR decoding requires multiple forward passes, which increases inference time and can be further optimized for contact-rich tasks requiring high control frequencies.

%------------------------------------------------------------------
\clearpage
\bibliographystyle{IEEEtran}
\bibliography{main}


\newpage
\twocolumn[  % Two-column with spanning title
\begin{@twocolumnfalse}
\begin{center}
{\LARGE \bf Appendix for \textit{CycleVLA}}
\end{center}
\vspace{1em}
\end{@twocolumnfalse}
]

\makeatletter
% \renewcommand{\@seccntformat}[1]{{\normalfont\csname the#1\endcsname.\quad}}
% \makeatother
\makeatletter
\renewcommand{\@seccntformat}[1]{\csname the#1\endcsname.\quad}
\makeatother
\setcounter{section}{0}
\renewcommand{\thesection}{\Alph{section}}
\renewcommand{\thesubsection}{\thesection.\arabic{subsection}}
\setcounter{secnumdepth}{2}
% reset hyperref's counter tracking
\renewcommand{\theHsection}{appendix.\Alph{section}}

%------------------------------------------------------------------
\section{Overview}
This Appendix includes: 1) details and human evaluation of our subtask decomposition pipeline, 2) implementation details of our MBR decoding strategy, 3) additional experiments and methodology details, 4) real-world system details, and 5) prompt details for the LLM subtask decomposition and VLM failure predictor and planner.

%------------------------------------------------------------------
\section{Details and Evaluation of Subtask Decomposition}\label{sec:Details and Evaluation of Subtask Decomposition}

\subsection{Movement Primitive Extraction}
\noindent\textbf{List of Movement Primitives.} Same as ECoT~\cite{ZawalskiCPMFL24}. Movement labels take the form:

\vspace{0.25em}
{\scriptsize
\noindent \texttt{move {[}forward/backward{]} {[}left/right{]} {[}up/down{]},\\
tilt {[}up/down{]},\\
rotate {[}clockwise/counterclockwise{]},\\
{[}close/open{]} gripper\\
}}

\noindent Components below threshold are omitted. If no movement is detected, the label is stop.

\vspace{0.5em}
\noindent\textbf{Sliding Window Classification.} Following~\cite{ZawalskiCPMFL24}, we classify movement primitives by computing the difference between robot states over a sliding window of 4 timesteps and thresholding each dimension to produce a discrete movement label.

\vspace{0.5em}
\noindent\textbf{Initial Thresholds.} For LIBERO, we convert axis-angle rotation to Euler angles and normalize gripper width from two finger positions to a single value in $[0, 1]$ using max finger distance of 0.04~m. We use separate thresholds for translation, rotation, and gripper:

\vspace{0.25em}
\noindent{LIBERO:} $[\tau_\text{trans}, \tau_\text{rot}, \tau_\text{grip}] = [0.02, 0.0075, 0.03]$

% \begin{itemize}
%     \item Bridge dataset: $[\tau_\text{trans}, \tau_\text{rot}, \tau_\text{grip}] = [0.03, 0.03, 0.03]$
%     \item LIBERO dataset: $[\tau_\text{trans}, \tau_\text{rot}, \tau_\text{grip}] = [0.02, 0.0075, 0.03]$
% \end{itemize}

\vspace{0.5em}
\noindent\textbf{Per-Trajectory Translation Threshold Optimization.} We optimize only the translation threshold per trajectory to minimize: 1) overlaps between translation and gripper movements, and 2) spurious stop labels. We grid search over $\tau_\text{trans} \in [\tau_\text{trans}^{\text{init}} - 0.01, \tau_\text{trans}^{\text{init}} + 0.01]$ with 50 steps, minimizing:
\begin{equation}
    \text{score} = 1.0 \times N_\text{overlaps} + 2.5 \times N_\text{stops}.
\end{equation}

% \noindent\textbf{Coordinate Transformations.} For LIBERO, we convert axis-angle rotation to Euler angles (roll, pitch, yaw) and normalize gripper width from two finger positions to a single value in $[0, 1]$ using max finger distance of 0.04m. (this is not necessary, but keep it here for reference)

\subsection{Gripper State Detection}
\noindent\textbf{Trajectory Segmentation.} We segment the trajectory into chunks by detecting transitions in gripper state $\gamma_t$. Each chunk is labeled with its dominant gripper value: close ($-1$), open ($+1$), or idle ($0$).

\vspace{0.5em}
\noindent\textbf{Multi-Threshold Voting.} To robustly detect gripper state changes, we run primitive extraction with three gripper thresholds: $[0.028, 0.03, 0.032]$. We extract the gripper dimension from each, average across thresholds, and round to obtain the final label $\in \{-1, 0, +1\}$ (close, idle, open). Note that this voting scheme is applied only to gripper state detection, not to movement primitives.

\vspace{0.5em}
\noindent\textbf{Filtering Abnormal States.} We apply a post-filter to remove isolated idle segments that are surrounded by longer sequences of consistent gripper actions. For an idle segment of length $L$, if $L_\text{left} + L_\text{right} > L$ where $L_\text{left}$ and $L_\text{right}$ are the lengths of consecutive consistent gripper actions ($-1$ or $+1$) on either side, we replace the idle segment with the surrounding value.

\subsection{Movement Primitive Downsampling for LLM-based Subtask Boundary Inference}
For trajectories whose length exceeds a fixed maximum threshold of 100 steps, we temporally downsample the movement primitive sequence before querying the LLM for subtask boundary inference. Given a trajectory of length $T > 100$, we uniformly sample indices with stride $\lceil T / 100 \rceil$ to preserve global temporal structure. The LLM predicts subtask boundaries over the downsampled index set, which are then projected back to the original trajectory by linear index mapping.

\subsection{Qualitative Results of Subtask-Decomposed Dataset}
We show qualitative examples of our constructed subtask-decomposed dataset with timestamp boundaries (one example from each LIBERO task suite) in Fig.~\ref{fig:demo_decomposed_subtask}.

\subsection{Human Evaluation}

\begin{table}[t]
\setlength{\abovecaptionskip}{0.1cm}
    \footnotesize
    \caption{\textbf{Human evaluation of subtask decomposition quality on LIBERO.}}
    % \vspace{-0.5em}
    \begin{adjustbox}{width=1.0\linewidth,center}
    \centering
    \begin{tabular}{lccccc}
    \toprule
    \textbf{Metric} & \textbf{Spatial} & \textbf{Object} & \textbf{Goal} & \textbf{Long} & \textbf{Average} \\ \noalign{\vskip 0.3ex}
    \hline \noalign{\vskip 1.0ex}
    Absolute Error (steps) $\downarrow$
        & 4.2 & 3.5 & 8.0 & 7.2 & 5.7 \\
    \noalign{\vskip 0.5ex}
    Relative Error (\%) $\downarrow$
        & 3.9 & 2.5 & 5.8 & 3.0 & 3.8 \\
    \bottomrule
    \end{tabular}
    \label{tab:subtask_decomp_human}
\end{adjustbox}
\vspace{-0.5em}
\end{table}

We evaluate the quality of the subtask-decomposed dataset via human annotation. For each LIBERO task suite, we randomly sample one demonstration from each of the 10 tasks (10 per suite, 40 total across four suites). For each sampled demonstration, human annotators inspect the subtask temporal boundaries predicted by the LLM and manually adjust the start and end timestamps when necessary.

We recruit five human evaluators. We report the mean absolute timestamp deviation between LLM-predicted and human-corrected boundaries and the relative boundary error, defined as the absolute timestamp deviation divided by the average subtask duration (i.e., trajectory length) for each subtask. Results in Table~\ref{tab:subtask_decomp_human} demonstrate the precision of our pipeline; tasks with less distinct gripper state boundaries and high degrees of rotation (e.g., LIBERO-Goal) are more challenging to decompose into subtask trajectories.

%------------------------------------------------------------------
\begin{table*}[t]
\setlength{\abovecaptionskip}{0.1cm}
    \footnotesize
    \caption{\textbf{Finetuning hyperparameters for OpenVLA-OFT-Diff on LIBERO.}}
    \begin{adjustbox}{width=1.0\linewidth,center}
    \centering
    \begin{tabular}{@{}p{3.5cm}p{12cm}@{}}
    \toprule
    \textbf{Hyperparameter} & \textbf{Value} \\
    \noalign{\vskip 0.3ex}
    \hline
    \noalign{\vskip 1.0ex}
  pretrained checkpoint & \texttt{openvla-7b}\\
  \noalign{\vskip 0.5ex}
    \# GPUs & 4 $\times$ NVIDIA A100 (40GB VRAM) \\
    \noalign{\vskip 0.5ex}
    learning rate (LR) & 5e-4 (decay to 5e-5 after 335K steps) \\
    \noalign{\vskip 0.5ex}
    effective batch size & 64 (2 per GPU, gradient accumulation of 8) \\
    \noalign{\vskip 0.5ex}
    \# train steps & 500K \\
    \noalign{\vskip 0.5ex}
    \# diffusion steps & 50 \\
    \noalign{\vskip 0.5ex}
    input images & 1 third-person camera image, 1 wrist-mounted camera image \\
    \noalign{\vskip 0.5ex}
    input image size & 224 $\times$ 224 px \\
    \noalign{\vskip 0.5ex}
    use observation history & no (use single-step inputs) \\
    \noalign{\vskip 0.5ex}
    LoRA rank & 32 \\
    \noalign{\vskip 0.5ex}
    action chunk size ($H$) & 8 steps (predict 8, execute all 8 open-loop at test time) \\
    \noalign{\vskip 0.5ex}
    action dimensions & 9 (7 robot + 1 stop + 1 progress) \\
    \noalign{\vskip 0.5ex}
    use proprio (robot state) & yes \\
    \noalign{\vskip 0.5ex}
    use FiLM & no \\
    \noalign{\vskip 0.5ex}
    \# trainable parameters & 313M total (111M LoRA adapter + 185M action head + 17M proprio projector) \\
    \noalign{\vskip 0.5ex}
    last-action oversampling & 8$\times$ \\
    \noalign{\vskip 0.5ex}
    image augmentations & 90\% random crops, color jitter: \newline
    \texttt{random\_resized\_crop=dict(scale=[0.9, 0.9], ratio=[1.0, 1.0]),} \newline
    \texttt{random\_brightness=[0.2], random\_contrast=[0.8, 1.2],} \newline
    \texttt{random\_saturation=[0.8, 1.2], random\_hue=[0.05]} \\
    \bottomrule
    \end{tabular}
    \label{tab:hyperparameters_training}
\end{adjustbox}
\vspace{-0.5em}
\end{table*}

\section{Implementation Details of MBR Decoding}\label{sec:Implementation Details of MBR Decoding}

We sample $N{=}8$ hypotheses by re-running stochastic decoding with different random seeds, which changes the noise sampling in the diffusion/flow matching action expert. We compute the $N \times N$ pairwise $L_2$ distance matrix over trajectory features, and select the hypothesis with minimum average distance.

\vspace{0.5em}
\noindent\textbf{Trajectory Features.} For each hypothesis, we accumulate the predicted translational and rotational deltas along the chunk to obtain a cumulative end-effector trajectory. Accumulating deltas rather than comparing them step-wise ensures that the distance reflects divergence in the resulting motion, so that small early deviations that compound over the chunk are weighted accordingly.

\vspace{0.5em}
\noindent\textbf{Pairwise Distance and Selection.} We compute the full $N \times N$ pairwise $L_2$ distance matrix over these features, and select the hypothesis minimizing the average distance to all hypotheses in $\mathcal{A}$, following Eq.~(4). The selected chunk is the medoid of the hypothesis
set under $d \circ \phi$, and is executed directly without further modification.

%Rather than selecting the hypothesis with minimum average distance (standard MBR), we use a density-based variant that identifies the densest region and selects its medoid.

% \vspace{0.5em}
% \noindent\textbf{Adaptive $r$-NN Density Estimation.} For each hypothesis, we compute its $r$-NN radius, the distance to its $r$-th nearest neighbor, as a local density estimate (smaller radius indicates higher density). We set $r$ adaptively as:
% \begin{equation}
% r = \max\left(2, \min\left(4, \lfloor\sqrt{N}\rfloor\right)\right),
% \end{equation}
% which yields $r{=}2$ for $N{=}8$.

% \vspace{0.5em}
% \noindent\textbf{Medoid Selection.} We identify the densest point (smallest $r$-NN radius) as the pocket center, then find the medoid within this pocket—the hypothesis that minimizes average distance to other pocket members. This medoid is selected as the final action chunk for execution.

%------------------------------------------------------------------
\begin{table*}[b]
\setlength{\abovecaptionskip}{0.1cm}
    \footnotesize
    \caption{\textbf{Runtime analysis of test-time scaling on LIBERO.} End-to-end runtime in seconds (s~$\downarrow$) for each component on different GPUs. Percentages indicate share of total inference time.}
    \begin{adjustbox}{width=0.85\linewidth,center}
    \centering
    \begin{tabular}{lcccccc}
    \toprule
    \textbf{GPU} & \textbf{VLM} & \textbf{Action Rollout} & \textbf{Action Sampling} & \textbf{MBR} & \textbf{Backtrack} & \textbf{Total} \\
    \noalign{\vskip 0.3ex}
    \hline 
    \noalign{\vskip 1.0ex}
    A10  & 12.9 {\textcolor{improvementcolor}{(6.0$\%$)}} & 147.6 {\textcolor{improvementcolor}{(68.6$\%$)}} & 47.9 {\textcolor{improvementcolor}{(22.2$\%$)}} & 0.003 {\textcolor{improvementcolor}{({<}0.1$\%$)}} & 6.9 {\textcolor{improvementcolor}{(3.2$\%$)}} & 215.3 \\
    \noalign{\vskip 0.5ex}
    A100 & 15.3 {\textcolor{improvementcolor}{(19.9$\%$)}} & 44.7 {\textcolor{improvementcolor}{(58.1$\%$)}} & 11.9 {\textcolor{improvementcolor}{(15.5$\%$)}} & 0.002 {\textcolor{improvementcolor}{({<}0.1$\%$)}} & 5.0 {\textcolor{improvementcolor}{(6.5$\%$)}} & 76.9 \\
    \bottomrule
    \end{tabular}
    \label{tab:runtime}
\end{adjustbox}
\vspace{-0.5em}
\end{table*}

\section{Additional Experiments and Methodology Details}\label{sec:Additional Experiments and Details}

\noindent\textbf{Robust Stop and Progress Signal Detection.} Since predicted stop ($s_t$) and progress ($p_t$) signals can be noisy, we apply a confirmation mechanism $\textsc{Confirm}(\cdot)$ before triggering subtask transitions or VLM checks. A high signal refers to either $p_t \geq \tau_p$ (for progress) or $s_t = 1$ (for stop).

We track three quantities: 1) if a high signal has been observed (\texttt{first\_seen}), 2) the count of consecutive high signals ($c_{\text{consec}}$), and 3) the number of low-signal steps since the last high signal ($c_{\text{gap}}$). A condition is confirmed if either:
\begin{equation}
c_{\text{consec}} \geq 2 \quad \text{or} \quad (\texttt{first\_seen} \land c_{\text{gap}} \geq 2),
\end{equation}
that is, two consecutive high signals, or a high signal that recurs after at least two low-signal steps. This filters isolated spurious predictions while remaining responsive to genuine transitions.

\vspace{0.5em}
\noindent\textbf{OpenVLA-OFT-Diff Baseline.} We finetune the OpenVLA-OFT-Diff baseline ourselves, since the original OpenVLA-OFT~\cite{kim2025fine} reports its diffusion variant only in a single-camera configuration without wrist images or proprioceptive state. We train a single monolithic policy across all LIBERO task suites, with a third-person camera image, a wrist camera image, and proprioceptive state as inputs, and with FiLM deactivated. The baseline uses the standard 7-dimensional action space and otherwise shares the training hyperparameters in Table~\ref{tab:hyperparameters_training}, without the stop and progress dimensions or last-action oversampling used by our method.

\vspace{0.5em}
\noindent\textbf{Training Hyperparameters.} Hyperparameters for finetuning the OpenVLA-OFT~\cite{kim2025fine} backbone with a diffusion-based action head (OpenVLA-OFT-Diff) on LIBERO are listed in Table~\ref{tab:hyperparameters_training}. For faster convergence, we decay the learning rate from 5e-4 to 5e-5 after 335K gradient steps. For the $\pi_{0.5}$~\cite{pi0_5_pi} backbone, we perform full finetuning initialized from the publicly released $\pi_{0.5}$-LIBERO checkpoint. Parameters are listed in Table~\ref{tab:hyperparameters_training_pi05}

% \subsection{Runtime Analysis of Test-Time Scaling}\label{sec:Runtime Analysis of Test-Time Scaling}
\vspace{0.5em}
\noindent\textbf{Runtime Analysis of Test-Time Scaling.} We analyze the contribution of each component in CycleVLA to inference-time overhead under test-time scaling with the OpenVLA-OFT-Diff backbone. For each task episode, we record the runtime of: 1) VLM failure prediction and planning (OpenAI API call, Tier-5 user), 2) VLA action inference and robot execution, 3) action sampling after backtracking, 4) MBR pairwise distance computation, and 5) backtracking execution. We compare runtime on NVIDIA A10 and A100 GPUs. From Table~\ref{tab:runtime}, standard action rollout remains the dominant bottleneck, while test-time scaling in CycleVLA adds only moderate overhead. With stronger GPUs, the relative proportion of VLM API latency increases, as action inference becomes faster. We consider this overhead acceptable, as test-time scaling methods in LLMs typically incur linear or near-linear runtime increases with the number of samples~\cite{snell_scale_llm, SunHZYQYWBZ24}.

We further note that the seemingly long end-to-end runtime is not primarily caused by CycleVLA itself. The dominant action rollout cost reflects the inference of the large-scale VLA backbone and the physical execution of the task, and is incurred by any policy regardless of failure correction. The components introduced by CycleVLA (VLM queries, action sampling, MBR computation, and backtracking) add only $\sim$30\% overhead on top of this baseline cost, with MBR selection itself being negligible ({<}0.1\%). CycleVLA's test-time scaling is thus efficient relative to the execution it corrects.

\vspace{0.5em}
\noindent\textbf{Inference Hyperparameters.} The VLM failure predictor is queried when subtask progress reaches $\tau_p = 0.9$. Each subtask allows up to $R=3$ retries before forcing completion. We sample $N=8$ hypotheses for MBR decoding with $L_2$ distance.

\begin{table*}[t]
\setlength{\abovecaptionskip}{0.1cm}
  \footnotesize
  \caption{\textbf{Finetuning hyperparameters for $\pi_{0.5}$ on LIBERO and real-robot experiments.}}
  \begin{adjustbox}{width=1.0\linewidth,center}
  \centering
  \begin{tabular}{@{}p{3.5cm}p{12cm}@{}}
  \toprule
  \textbf{Hyperparameter} & \textbf{Value} \\
  \noalign{\vskip 0.3ex}
  \hline
  \noalign{\vskip 1.0ex}
  pretrained checkpoint & \texttt{pi05\_libero}\\
  \noalign{\vskip 0.5ex}
  \# GPUs & 8 $\times$ NVIDIA A100 (40GB VRAM) \\
  \noalign{\vskip 0.5ex}
  finetuning method & full (all ${\sim}$3.2B parameters trainable) \\
  \noalign{\vskip 0.5ex}
  optimizer & AdamW ($\beta_1{=}0.9$, $\beta_2{=}0.95$, $\epsilon{=}10^{-8}$, weight decay
$10^{-10}$, grad-norm clip $1.0$) \\
  \noalign{\vskip 0.5ex}
  learning rate (LR) & 5e-5 (constant after 10K linear warmup steps) \\
  \noalign{\vskip 0.5ex}
  effective batch size & 128 (16 per GPU $\times$ 8 GPUs, FSDP sharding) \\
  \noalign{\vskip 0.5ex}
  \# train steps & 30K \\
  \noalign{\vskip 0.5ex}
  \# denoising steps (inference) & 10 (flow-matching) \\
  \noalign{\vskip 0.5ex}
  input images & 1 third-person camera image, 1 wrist-mounted camera image \\
  \noalign{\vskip 0.5ex}
  input image size & 224 $\times$ 224 px \\
  \noalign{\vskip 0.5ex}
  use observation history & no (single-step inputs) \\
  \noalign{\vskip 0.5ex}
  action chunk size ($H$) & 10 steps (predict 10, execute 5 open-loop at test time) \\
  \noalign{\vskip 0.5ex}
  action dimensions & 9 (7 robot + 1 stop + 1 progress) \\
  \noalign{\vskip 0.5ex}
  use proprio (robot state) & yes (continuous input) \\
  \noalign{\vskip 0.5ex}
  EMA decay & 0.999 \\
  \noalign{\vskip 0.5ex}
    last-action oversampling & 8$\times$ \\
    \noalign{\vskip 0.5ex}
  image augmentations & Non-wrist: 95\% random crop + resize back, $\pm5^{\circ}$ rotation, color
jitter (brightness 0.3, contrast 0.4, saturation 0.5). Wrist: color jitter only \\
  \bottomrule
  \end{tabular}
  \label{tab:hyperparameters_training_pi05}
\end{adjustbox}
\vspace{-0.5em}
\end{table*}

\begin{figure*}[t]
  \centering
  \footnotesize
  \setlength{\abovecaptionskip}{0.1cm}
  \includegraphics[width=1\linewidth]{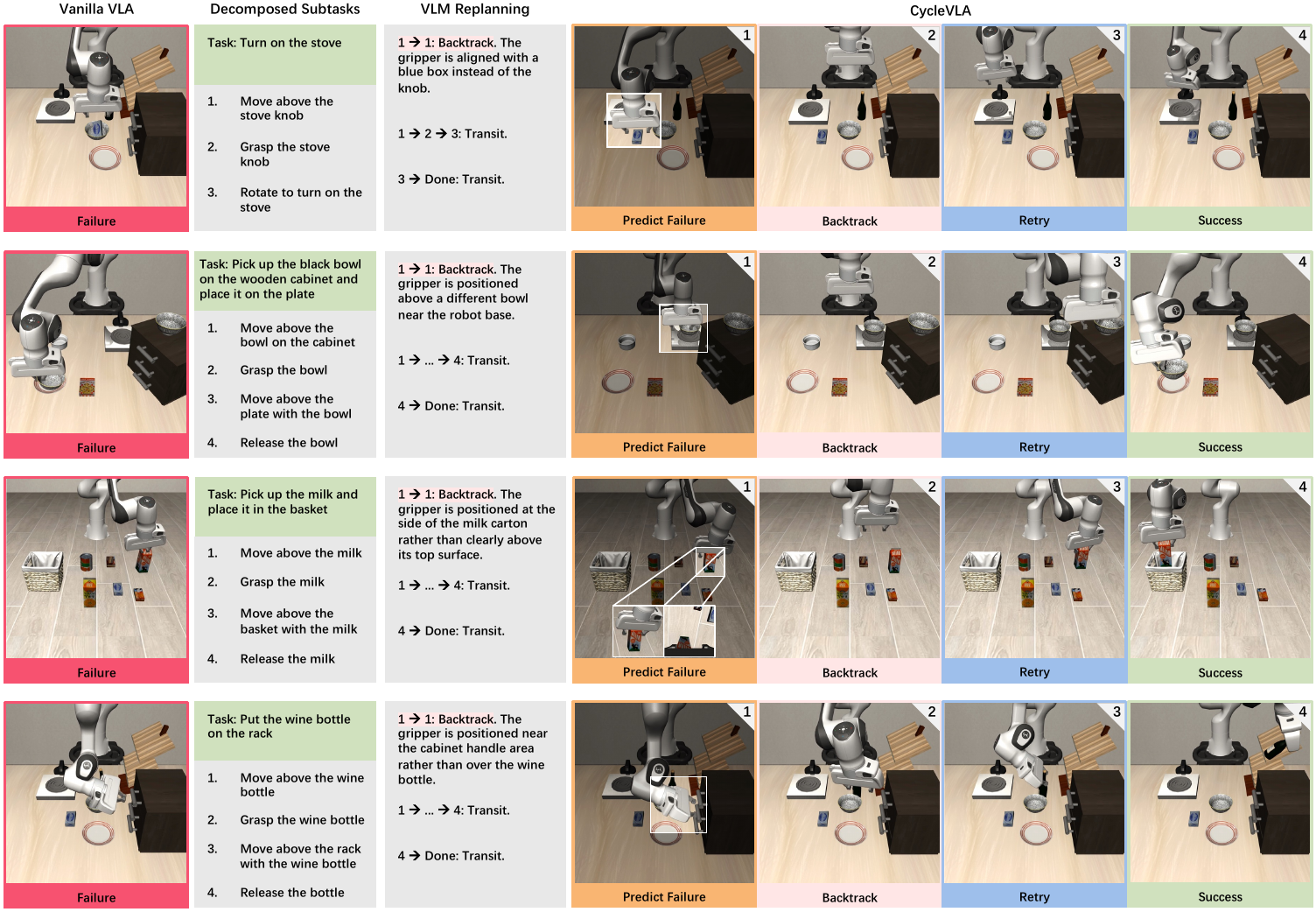}
  \vspace{-3mm} \caption{\textbf{Additional examples of CycleVLA.}}
  \label{fig:more_demo}
  % \vspace{-1.0em} 
\end{figure*}

\vspace{0.5em}
\noindent\textbf{More Qualitative Results.} We show more qualitative examples in Fig.~\ref{fig:more_demo}.

%------------------------------------------------------------------
\section{Additional Details of Real System}\label{sec:Additional Details of Real System}
\noindent\textbf{Control and Execution.} We use an AgileX PiPER 6-DOF robot arm with a single gripper, controlled via CAN bus. As in simulation, the system operates in end-effector space: at each 20\,Hz control step, the policy outputs a 9-dimensional action $a_t = [\Delta x_t, \Delta y_t, \Delta z_t, \Delta u_t, \Delta v_t, \Delta w_t, \gamma_t, s_t, p_t]^\top$, where $(\Delta x_t, \Delta y_t, \Delta z_t)$ and $(\Delta u_t, \Delta v_t, \Delta w_t)$ denote translational and rotational end-effector displacements, $\gamma_t$ is the gripper command, $s_t$ is the stop signal, and $p_t$ is the subtask progress signal. The displacements are applied additively in Euler space ($\text{target} = \text{current} + \Delta$, with orientation wrapped to $(-\pi, \pi]$), and the firmware solves inverse kinematics internally via its Cartesian \texttt{EndPoseCtrl} interface. The policy produces action chunks of size $H=10$, of which the first 5 actions are executed open-loop before requerying.

\begin{figure}[t]
  \centering
  \footnotesize
  \setlength{\abovecaptionskip}{0.1cm}
  \includegraphics[width=1\linewidth]{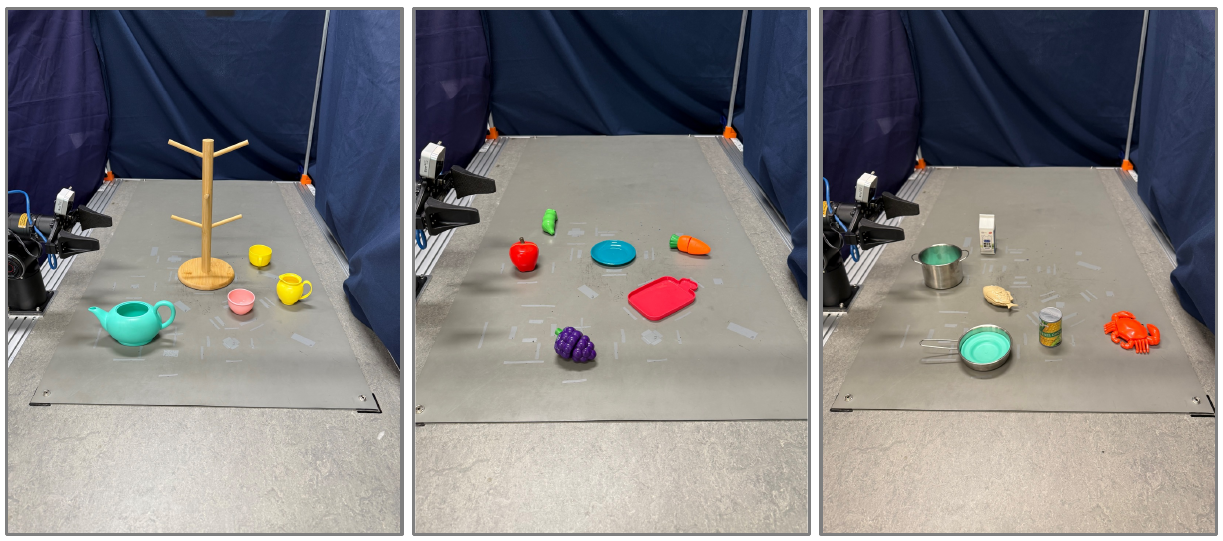}
    \vspace{-3mm} \caption{\textbf{Example initializations of the three real-world tasks.} Left: teapot hanging. Middle: fruit sorting. Right: cookware packing. Object positions are randomized within $\pm$30\,cm and orientations within $\pm$60$^\circ$ across trials.}
  \label{fig:init}
  \vspace{-1.0em} 
\end{figure}

\vspace{0.5em}
\noindent\textbf{Teleoperation and Gripper Handling.} During demonstrations, we teleoperate the PiPER arm using a leader--follower setup: the operator physically moves a leader arm by hand, and the follower mirrors its motion through a firmware-level CAN linkage. Since the operator continuously squeezes and releases the leader gripper, the recorded gripper trajectory is inherently continuous, unlike simulation environments where gripper actions are toggled. We retain this continuous gripper representation throughout training (i.e., $\gamma_t$ is continuous rather than binary on the real system). At deployment, however, the policy's continuous gripper output is binarized via an opening-width threshold ($\gamma_t \leq \tau_{\text{close}} \rightarrow$ full close; $\gamma_t > \tau_{\text{close}} \rightarrow$ position-control tracking) to ensure firm grasps on physical objects. We use $\tau_{\text{close}} = 6.0$\,cm by default (fruit sorting and cookware packing) and $\tau_{\text{close}} = 8.25$\,cm for teapot hanging, where the wider threshold triggers a firm close earlier to hold the teapot handle securely.

\vspace{0.5em}
\noindent\textbf{Backtracking.} The system records joint positions at every control step during the forward rollout. When the VLM failure predictor triggers a backtrack, the arm rewinds deterministically by replaying the recorded joints in reverse via \texttt{JointCtrl}, then resumes in end-effector mode for the retry.

\vspace{0.5em}
\noindent\textbf{Task Setup and Initialization.} Fig.~\ref{fig:init} shows an example initialization for each of the three real-world tasks.

\vspace{0.5em}
\noindent\textbf{Training.} Real-robot finetuning of $\pi_{0.5}$ uses the same hyperparameters as in Table~\ref{tab:hyperparameters_training_pi05}, with demonstrations processed as described below.

\vspace{0.5em}
\noindent\textbf{Demonstration Processing.} We collect 100 demonstrations per task at 20\,Hz using the leader--follower teleoperation setup. During recording, the operator presses a key to mark subtask boundaries; the per-frame subtask index is used offline to segment each episode and assign per-subtask language instructions. Two Intel RealSense D405 cameras (one wrist-mounted, one third-person) capture $640 \times 480$ RGB images, which are resized with aspect-preserving padding to $224 \times 224$ for the policy. Raw demonstrations undergo DROID-style\cite{khazatsky2024droid} no-op filtering per subtask: frames whose motion from the previous \emph{kept} frame falls below threshold (position $< 0.5$\,mm, orientation $< 5$\,mrad, gripper $< 10^{-3}$) are dropped, while the first and last frames of each subtask are always retained. End-effector delta actions are computed as consecutive-frame differences with orientation deltas wrapped to $(-\pi, \pi]$; the gripper command $\gamma_t$ is the continuous absolute gripper position rather than a delta.

Each subtask's filtered frames receive a fractional progress signal $p_t$ that rises from $0.1$ to $0.9$ over the subtask body, discretized in $0.1$ bins. The tail is oversampled: the last frame is repeated 8 times with $s_t = 1$ and $p_t = 1.0$, matching the simulation data pipeline. The 8-dimensional proprioceptive state is $[\text{EEF}(6), \gamma_t, -\gamma_t]$: since the PiPER has a single gripper DOF while the pretrained model expects a two-finger gripper state from simulation, the gripper position fills both slots as $[\gamma_t, -\gamma_t]$ to match the expected input dimensionality. Normalization statistics absorb the scale difference between the real and simulated grippers.

\begin{figure*}[t]
  \centering
  \footnotesize
  \setlength{\abovecaptionskip}{0.1cm}
  \includegraphics[width=1\linewidth]{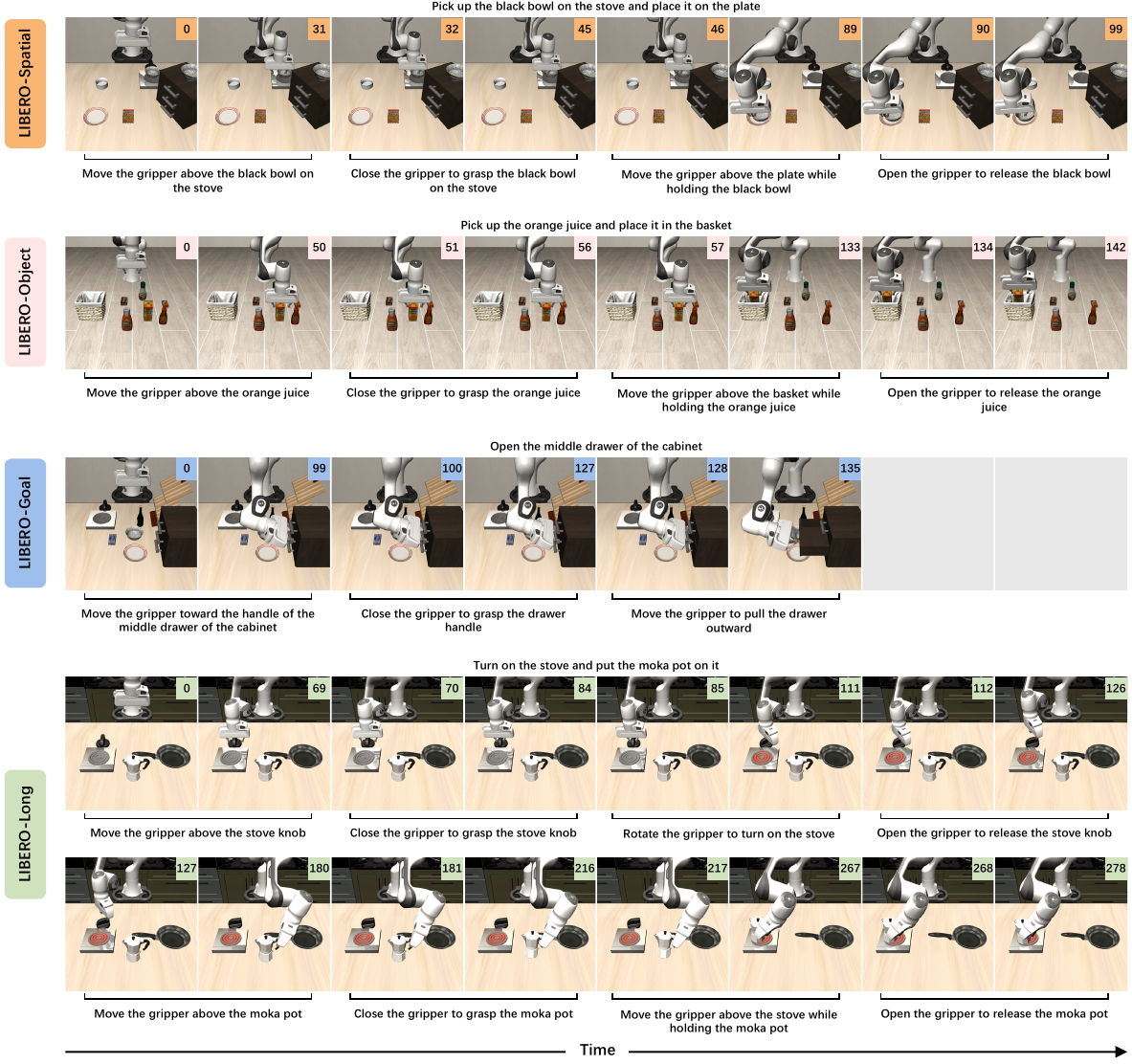}
  \vspace{-3mm} \caption{\textbf{Qualitative examples of subtask-decomposed dataset.}}
  \label{fig:demo_decomposed_subtask}
  % \vspace{-1.0em} 
\end{figure*}

%------------------------------------------------------------------
\clearpage
\onecolumn
\section{Prompt Details of Subtask Decomposition}\label{sec:Prompt Details of Subtask Decomposition}
We show exact prompts for LLMs to propose subtasks and infer their timestamp boundaries based on movement primitive sequence.

\begin{tcolorbox}[mypromptstyle]
\prompttitle{Subtask Proposal Prompt.}
\begin{flushleft}
\setlength{\parindent}{0pt}
\setlength{\leftskip}{1em}
\setlength{\rightskip}{1em}
Input:\\
1. Task: \{language\_instruction\}\\
\vspace{1em}
Instruction:\\
You are given a high-level robotic task description. Your job is to decompose this task into a minimal set of formal subtasks that a robot must perform to complete the task.\\
\vspace{1em}
Rules:\\
1. Minimal Subtask Decomposition\\
\hspace{1em}- Focus on the necessary and sufficient actions to complete the task.\\
\hspace{1em}- Do not over-decompose — prefer atomic but essential steps.\\
\hspace{1em}\parbox[t]{\dimexpr\linewidth-3em}{\hangindent=0.6em\hangafter=1 - Ask: What does the robot must do to succeed, regardless of variations in execution?}\\
2. Object-Centric Reasoning\\
\hspace{1em}- Use the presence and spatial arrangement of objects as key indicators.\\
\hspace{1em}\parbox[t]{\dimexpr\linewidth-3em}{\hangindent=0.6em\hangafter=1 - Mention relative spatial cues (e.g., above the drawer handle) if implied in the task.}\\
3. Skillset Usage and Formal Language\\
\hspace{1em}- Each subtask must begin with one of these actions (verbs):\\
\hspace{2em}1) "Move the gripper ..."\\
\hspace{2em}2) "Rotate the gripper ..."\\
\hspace{2em}3) "Open the gripper ..."\\
\hspace{2em}4) "Close the gripper ..."\\
\hspace{1em}- Keep language precise, formal, and robotic.\\
\hspace{1em}- As a rule of thumb:\\
\hspace{2em}1) Move/Rotate is often followed by Close to grasp.\\
\hspace{2em}2) Grasped objects are then moved, followed by Open to release.\\
\vspace{1em}
Output Format:\\
Subtasks: ["<subtask\_1>", "<subtask\_2>", ...]\\
Reasoning: "<Explain how and why you chose each subtask based on the instruction>"
\end{flushleft}

\prompttitle{Subtask Timestamp Boundary Inference Prompt.}
\begin{flushleft}
\setlength{\parindent}{0pt}
\setlength{\leftskip}{1em}
\setlength{\rightskip}{1em}
You are an expert reinforcement learning researcher. You have trained an optimal policy to control a robotic arm, which successfully completed a task as specified in natural language. The robot executed a sequence of actions to complete this task. Each action is recorded as a step in a trajectory.\\
\vspace{1em}
Input:\\
1. Task: \{language\_instruction\}\\
2. Subtasks list: \{subtasks\}\\
3. Trajectory features: Each entry in the dictionary corresponds to a single step on the trajectory and describes the move that is about to be executed.\\
\hspace{1em}trajectory\_features = \{0: "<move\_primitive>", 1: "<move\_primitive>", ...\}\\
\vspace{1em}
Instruction:\\
Decompose the entire trajectory into subtasks and assign a start and end step index to each subtask. The goal is to produce a mapping of the form:\\
Labeled\_dict = \{"subtask\_1": [start\_idx, end\_idx], "subtask\_2": [start\_idx, end\_idx], ...\}\\
\vspace{1em}
Rules:\\
1. Noisy Labels\\
\hspace{1em}- The trajectory data contains noise. Apply the following rules carefully:\\
\hspace{1em}- "stop" labels often appear even when the robot is still moving.\\
\hspace{1em}- Treat "stop" as movement if it appears between meaningful motion steps.\\
\hspace{1em}- Do NOT segment a new subtask just because you see a "stop".\\
\hspace{1em}\parbox[t]{\dimexpr\linewidth-3em}{\hangindent=0.6em\hangafter=1 - "open/close gripper" labels that persist for very short durations may be noise. However, gripper actions are generally more reliable than motion primitives.}\\[0.4em]
2. Robust Labeling Strategies\\
\hspace{1em}\parbox[t]{\dimexpr\linewidth-3em}{\hangindent=0.6em\hangafter=1 - Brief, short, inconsistent movement descriptions that conflict with prior and subsequent steps are likely to be noise.}\\
\hspace{1em}\parbox[t]{\dimexpr\linewidth-3em}{\hangindent=0.6em\hangafter=1 - Cross-reference other movement labels to decide whether a short-duration label is meaningful or just noise.}\\
3. Exhaustive Coverage\\
\hspace{1em}\parbox[t]{\dimexpr\linewidth-3em}{\hangindent=0.6em\hangafter=1 - You must not skip any steps. Every step in the trajectory should be assigned to exactly one subtask.}\\
\hspace{1em}\parbox[t]{\dimexpr\linewidth-3em}{\hangindent=0.6em\hangafter=1 - There should be no gaps in the index ranges. Subtasks must be labeled sequentially and cover the entire range of the trajectory indices.}\\[0.4em]
\vspace{1em}
Output Format:\\
Labeled\_dict: \{"subtask\_1": [start\_idx, end\_idx], "subtask\_2": [start\_idx, end\_idx], ...\}\\
Reasoning: "<Explain your logic for identifying and segmenting each subtask>"
\end{flushleft}

\end{tcolorbox}

%------------------------------------------------------------------
\vspace{1.0em}
\section{Prompt Details of Failure Predictor and Planner}\label{sec:Prompt Details of Failure Predictor and Planner}
We show exact prompts for VLMs to predict failure and plan recovery at subtask boundaries.

\begin{tcolorbox}[mypromptstyle]
\prompttitle{VLM Failure Prediction and Planning Prompt.}
\begin{flushleft}
\setlength{\parindent}{0pt}
\setlength{\leftskip}{1em}
\setlength{\rightskip}{1em}
You are an expert robot behavior annotator. Decide what the robot should do next given it is \textasciitilde90\% through the current subtask. Your job is to FORECAST whether the current subtask will likely succeed if we continue without corrective repositioning.\\
\vspace{1em}
Input:\\
1. Task instruction: \{language\_instruction\}\\
2. Subtask list: \{subtasks\}\\
3. Current subtask: \{current\_subtask\}\\
4. Visual inputs (two synchronized views):\\
\hspace{1em}- FRONT: third-person view (global alignment, object identity, spatial relations)\\
\hspace{1em}- WRIST: close-up gripper view (detailed contact, local geometry, physical affordances)\\
\vspace{1em}
Decision Rule (forecasting at \textasciitilde90\%):\\
\hspace{1em}\parbox[t]{\dimexpr\linewidth-3em}{\hangindent=0.6em\hangafter=1 - Default to transit when success appears reasonably likely within the next few actions without corrective repositioning.}\\
\hspace{1em}\parbox[t]{\dimexpr\linewidth-3em}{\hangindent=0.6em\hangafter=1 - Choose backtrack if strong, unambiguous visual evidence indicates that the subtask will fail without repositioning.}\\
\vspace{1em}
View-Specific Fusion Instruction:\\
\hspace{1em}\parbox[t]{\dimexpr\linewidth-3em}{\hangindent=0.6em\hangafter=1 - FRONT view provides global context: object identity, pose, global alignment, reachability, and path clearance.}\\
\hspace{1em}\parbox[t]{\dimexpr\linewidth-3em}{\hangindent=0.6em\hangafter=1 - WRIST view provides local interaction cues: gripper orientation, contact points, slip, stability, and detailed positioning relative to affordances.}\\[0.4em]
\hspace{1em}\parbox[t]{\dimexpr\linewidth-3em}{\hangindent=0.6em\hangafter=1 - Combine both views to reason about functional success: whether the current configuration supports the intended physical interaction (e.g., grasping, pulling, pushing).}\\[0.4em]
\hspace{1em}- FRONT dominates for global spatial reasoning and goal reachability.\\
\hspace{1em}- WRIST dominates for local contact accuracy and grasp quality.\\
\vspace{1em}
Affordance Reasoning Guidance:\\
\hspace{1em}\parbox[t]{\dimexpr\linewidth-3em}{\hangindent=0.6em\hangafter=1 - Evaluate whether the gripper's pose is consistent with the object's intended use:}\\
\hspace{2em}\parbox[t]{\dimexpr\linewidth-4em}{\hangindent=1.2em\hangafter=1 1) For a bowl: grasping the edge or rim is acceptable and often intended for lifting.}\\
\hspace{2em}\parbox[t]{\dimexpr\linewidth-4em}{\hangindent=1.2em\hangafter=1 2) For a drawer: alignment with the handle is the key indicator of readiness.}\\
\hspace{2em}\parbox[t]{\dimexpr\linewidth-4em}{\hangindent=1.2em\hangafter=1 3) For a push or pull action: confirm direction and surface contact match the required motion.}\\
\hspace{1em}\parbox[t]{\dimexpr\linewidth-3em}{\hangindent=0.6em\hangafter=1 - Do not penalize partial or asymmetric contacts if they serve a valid affordance and appear stable.}\\
\vspace{1em}
Wrong Object or Wrong Subtask Detection:\\
\hspace{1em}\parbox[t]{\dimexpr\linewidth-3em}{\hangindent=0.6em\hangafter=1 - In addition to misalignment, detect late-stage "silent failures" involving wrong object engagement or wrong subtask execution.}\\[0.4em]
\hspace{1em}\parbox[t]{\dimexpr\linewidth-3em}{\hangindent=0.6em\hangafter=1 - If visual evidence indicates the gripper is interacting with an unintended object, target, or affordance (e.g., lifting or contacting a distractor, manipulating the wrong receptacle, or committing to a different subtask's goal), output type: backtrack.}\\[0.4em]
\hspace{1em}\parbox[t]{\dimexpr\linewidth-3em}{\hangindent=0.6em\hangafter=1 - Set next\_subtask to the earliest subtask that restores correct target selection and preconditions (typically a reach, align, or target-identification step, NOT a trivial open/close gripper).}\\[0.4em]
\vspace{1em}
Backtracking Target:\\
\hspace{1em}\parbox[t]{\dimexpr\linewidth-3em}{\hangindent=0.6em\hangafter=1 - Backtrack to the earliest subtask that restores the missing precondition (typically a positioning or alignment step that enables correct affordance engagement).}\\[0.4em]
\vspace{1em}
Output Format:\\
next\_subtask: <exact subtask from subtasks list>\\
type: <transit / backtrack>\\
reason: <explain in a concise paragraph, justifying the decision>\\
\vspace{0.5em}
front\_view\_evidence:\\
\hspace{1em}- <concise observable cue 1>\\
\hspace{1em}- <concise observable cue 2>\\
\hspace{1em}- <concise observable cue 3>\\
\hspace{1em}- <concise observable cue 4>\\
\vspace{0.5em}
wrist\_view\_evidence:\\
\hspace{1em}- <concise observable cue 1>\\
\hspace{1em}- <concise observable cue 2>\\
\hspace{1em}- <concise observable cue 3>\\
\hspace{1em}- <concise observable cue 4>\\
\vspace{0.5em}
assessment:\\
\hspace{1em}- success\_likelihood: <high | medium | low>\\
\hspace{1em}- key\_risks: <comma-separated brief phrases>\\
\hspace{1em}- view\_agreement: <agree | partial | disagree>; <short phrase on which view dominates and why>\\
\hspace{1em}- decision\_basis: <short phrase linking likelihood + dominant cues to decision>\\
\vspace{1em}
Constraints:\\
\hspace{1em}- Focus strictly on observable visual and physical evidence.\\
\hspace{1em}- Keep each bullet concise ($\leq$12 words).\\
\hspace{1em}- Use exact strings from subtasks for next\_subtask.\\
\hspace{1em}- type must be either transit or backtrack.\\
\hspace{1em}- Return only the specified fields; no extra commentary.
\end{flushleft}

\end{tcolorbox}

%------------------------------------------------------------------

\end{document}